\documentclass[10pt,twocolumn,letterpaper]{article}

\usepackage[pagenumbers]{misc/cvpr}

\usepackage{graphicx}
\usepackage{amsmath}
\usepackage{amssymb}
\usepackage{booktabs}
\usepackage{misc/mo}
\usepackage{tabularx}
\usepackage{inconsolata}
\usepackage[accsupp]{axessibility}

\def\cvprPaperID{1219}
\def\confName{CVPR}
\def\confYear{2023}

\usepackage{amsmath,amsfonts,bm}

\def\eqref#1{equation~\ref{#1}}
\def\1{\bm{1}}

\DeclareMathAlphabet{\mathsfit}{\encodingdefault}{\sfdefault}{m}{sl}
\SetMathAlphabet{\mathsfit}{bold}{\encodingdefault}{\sfdefault}{bx}{n}

\DeclareMathOperator*{\argmax}{arg\,max}

\newcommand{\fewshot}[2]{ #1\textsubscript{\tiny\textcolor{Gray}{~#2}}}

\title{Learning Visual Representations via Language-Guided Sampling}
\author{
    Mohamed El Banani \quad Karan Desai \quad Justin Johnson \\
    University of Michigan \\
    {\tt \small \{mbanani,kdexd,justincj\}@umich.edu}
}

\begin{document}

\maketitle

\global\csname @topnum\endcsname 0
\global\csname @botnum\endcsname 0

\begin{figure}
  \centering
  \includegraphics[width=0.9\linewidth]{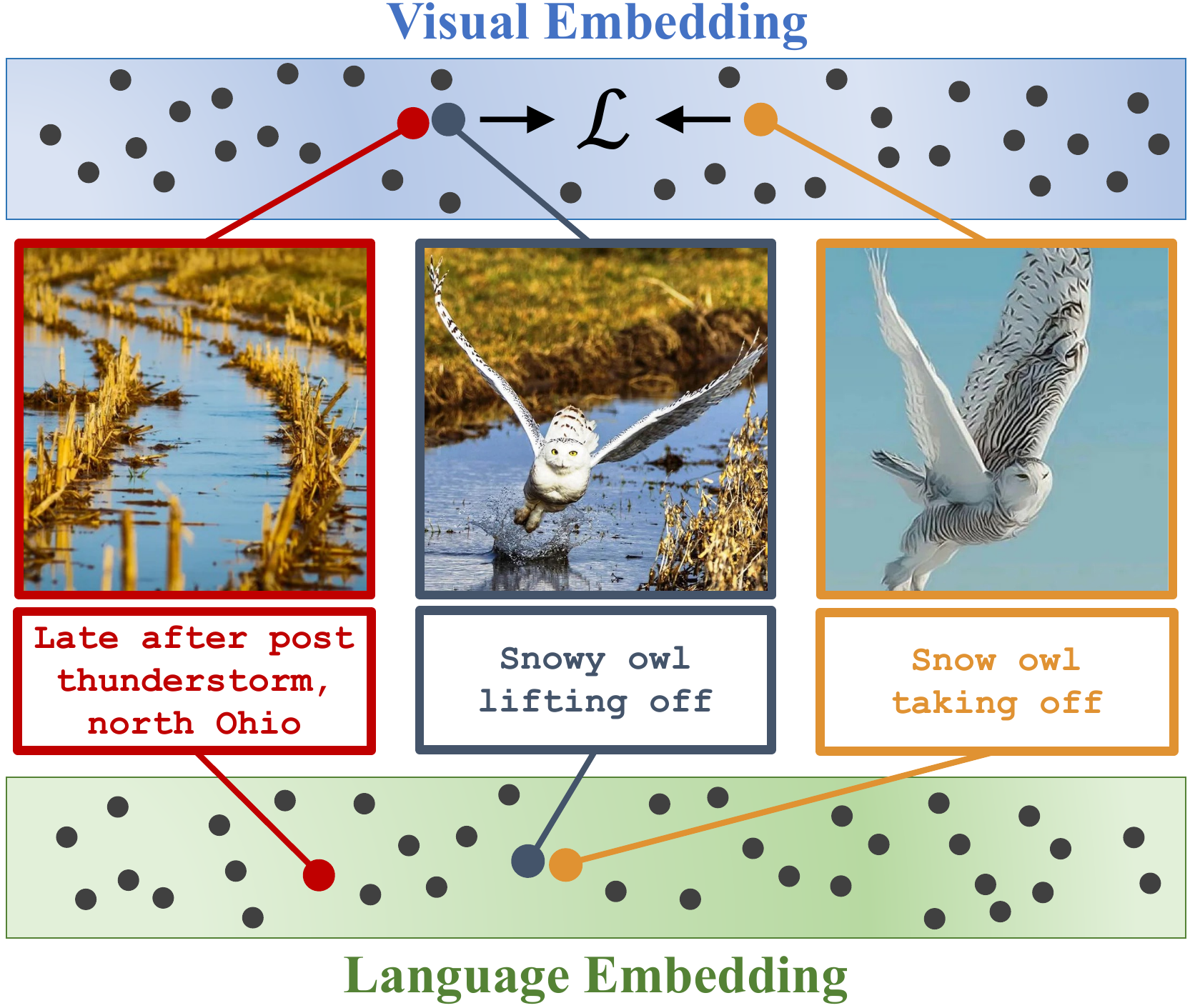}
  \caption{Language allows us to find conceptually similar image pairs even if they are visually dissimilar. We use those pairs for contrastive learning to learn generalizable visual features.}
  \label{fig:teaser}
\end{figure}

\begin{abstract}
Although an object may appear in numerous contexts, we often describe it in a limited number of ways.  
Language allows us to abstract away visual variation to represent and communicate concepts.  
Building on this intuition, we propose an alternative approach to visual representation learning: using language similarity to sample semantically similar image pairs for contrastive learning.  
Our approach diverges from image-based contrastive learning by sampling view pairs using language similarity instead of hand-crafted augmentations or learned clusters. 
Our approach also differs from image-text contrastive learning by relying on pre-trained language models to guide the learning rather than directly minimizing a cross-modal loss.  
Through a series of experiments, we show that language-guided learning yields better features than image-based and image-text representation learning approaches.

\end{abstract}

\section{Introduction}
\label{sec:intro}
Consider the images in~\cref{fig:teaser}, is the center image more similar to its left or right neighbor? 
Despite the difference in background and pose, it is clear that the right pair captures the same concept: a flying snow owl. 
Nevertheless, a self-supervised image model will judge the left pair as more similar. 
Human perception and language abstract away appearance differences to capture conceptual similarity rather than just visual similarity. 
Ideally, we could learn visual features that capture conceptual similarity and generalize effectively to other visual tasks. 
In this work, we show how language can be a proxy for conceptual similarity; allowing us to sample better pairs for contrastive learning and train more generalizable visual models.

Image-only contrastive learning uses visual similarity as a proxy for conceptual similarity.
This is based on the observation that discriminative approaches can discover inter-class similarity--\eg, cheetahs are similar to lions-- without requiring explicit annotations~\cite{wu2018unsupervised}. 
The core idea is to train a discriminative model where each instance is treated as a separate class, and the model is trained to map augmented views of the same image to similar features~\cite{chen2020mocov2,chen2020big,chen2020simple,chen2021exploring,wu2018unsupervised}.
While successful, instance discrimination ignores the similarity between different instances as it assumes all other images are unrelated.
Later work focused on inter-image relationships by estimating clusters~\cite{asano2019sela,caron2018deepcluster,caron2020unsupervised} or finding nearest neighbors~\cite{dwibedi2021nnclr}. 
However, those relationships are estimated using visual embeddings; resulting in visually, rather than conceptually, similar pairs.

Language similarity is a strong proxy for semantic relationships. 
Consider the example in \cref{fig:teaser}; images that depict the same concept are often described similarly. 
Radford~\etal~\cite{radford2021learning} propose language-image contrastive learning by mapping images and text to a shared representation space and achieve impressive generalization capabilities. 
However, it is unclear whether forcing models to map onto a shared space is optimal for visual learning. 
Although linguistic and visual similarity might align for similar instances, it is unclear whether all distances in one space should map exactly to the other. 
Instead of learning a joint vision-and-language representations, we argue that it is better to use linguistic similarity to guide visual learning.

\begin{figure*}
  \centering
  \includegraphics[width=\linewidth]{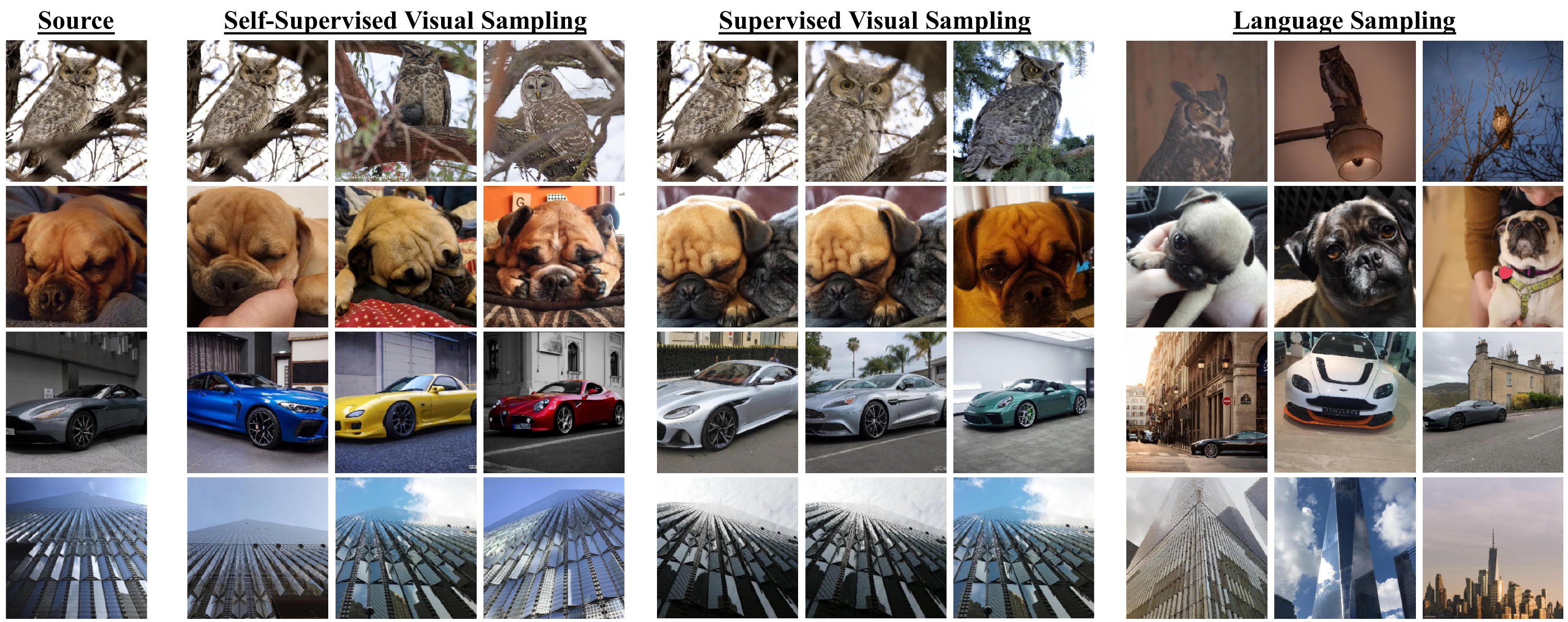}
  \caption{\textbf{Language sampling yields semantically-similar and visually-diverse image pairs.} 
  We sample the three nearest neighbors using a self-supervised visual model~\cite{chen2020big}, an ImageNet supervised model~\cite{dosovitskiy2021vit}, and a self-supervised language model~\cite{reimers2019sbert}. While visual sampling yields visually similar pairs, language sampling yields semantically relevant and visually diverse images. We argue that the combination of semantic consistency and visual diversity are better for learning generalizable features. 
}
  \label{fig:sampled_pairs}
\end{figure*}

To this end, we propose \emph{language-guided contrastive learning}:
a simple adaptation to contrastive learning that uses language models to find conceptually-similar image pairs for visual learning.
Our approach is motivated by the observation that language models, despite never training on visual data, can still be used to sample caption pairs that belong to conceptually similar images, as seen in~\cref{fig:sampled_pairs}.
Such sampled images exhibit desirable variations in pose, lightning, and context which are very different from hand-crafted augmentations which can be ill-suited to downstream tasks~\cite{xiao2020should} or too focused on background textures~\cite{selvaraju2021cast}. 
We use the sampled pairs instead of image augmentations within standard self-supervised visual learning approaches such as SimCLR~\cite{chen2020simple}, SimSiam~\cite{chen2021exploring}, and SLIP~\cite{mu2021slip}.
Our approach departs from image-only contrastive learning by relying on conceptually-similar image pairs rather than visually similar augmentations or cluster-assignment.
We also depart from image-text pre-training by allowing the model to be guided by language similarity rather than learning a joint embedding space. 

We conduct a series of controlled experiments to analyze our approach and compare it to commonly used representation learning paradigms on generalization to downstream classification tasks.
In controlled settings, our approach outperforms all baselines on linear probe and fewshot classification on a range of downstream classification datasets.  
Our analysis suggests that while learning multi-modal joint embeddings can result in good representations, it is better to use one modality to guide the training of the other.
Furthermore, we find that our approach is robust to the specific choice of sampling strategy or language model.
Our code and pre-trained models are available at \url{https://github.com/mbanani/lgssl}.

\section{Related Work}
\label{sec:related}

\nsparagraph{Visual Representation Learning} aims to learn visual embedding spaces that capture semantics, with a typical focus on learning from scalable data sources.
Broadly speaking, there are two general approaches: generative and discriminative. 
Generative approaches hypothesize that a model that can capture the image distribution will learn semantically relevant features~\cite{gidaris2018unsupervised,doersch2015unsupervised,zhang2016colorful,oord2018representation,he2022masked,vincent2008extracting}. 
In contrast, discriminative approaches posit that differentiating between images will give rise to better features. 
This idea can be traced by to early work on metric learning~\cite{chopra2005learning} and dimensionality reduction~\cite{hadsell2006dimensionality}, and is clearly seen for supervised classification models~\cite{sharif2014cnn}.
More recently, Wu~\etal~\cite{wu2018unsupervised} proposed treating each image as a separate class and using augmented images as class instances to relieve the need for human annotation. 
This was followed by papers that simplified this approach~\cite{chen2020simple,chen2020big,he2019moco,chen2020mocov2} and proposed non-contrastive variants~\cite{chen2021exploring,grill2020bootstrap}.
While those approaches have been successful, the utility of augmentation-based self-supervised learning has been questioned~\cite{newell2020useful,xiao2020should} with follow-up work proposing the use of objectness~\cite{peng2022crafting,mishra2021object} and saliency~\cite{selvaraju2021cast} to alleviate some of those concerns.
While we share the goal of visual representation learning, we question the reliance on image augmentations for training and propose using language models to learn for conceptually-similar images.

\nsparagraph{Language-supervised vision pre-training} aims to learn visual representations from language data.
Early work of Li~\etal~\cite{li2017ngrams} trained n-gram models using YFCC~\cite{yfcc100m} images and user-tag metadata.
While some works learn joint vision-and-language representations for tasks like
visual question answering~\cite{antol2015vqa,zhu2016visual7w,goyal2017making,hudson2019gqa},
visual reasoning~\cite{suhr2019corpus,zellers2019recognition,kazemzadeh2014referitgame},
and retrieval~\cite{young2014image,park2022normalized}, we are interested in using language to learn better visual representations~\cite{desai2021virtex,desai2021virtex,sariyildiz2020learning,radford2021learning,stroud2020learning}.
Early works used language modeling as a pretext task for visual learning~\cite{sariyildiz2020learning,desai2021virtex},
but contrastive approaches quickly gained more popularity due to their relative simplicity and generalization capabilities~\cite{radford2021learning,jia2021scaling}.
Follow-up work extended the contrastive formulation to learn dense features~\cite{yao2022filip,xu2022groupvit} or used additional self-supervised losses to improve performance and data efficiency~\cite{li2022declip,cui2022democratizing,mu2021slip,lee2022uniclip}. 
While we share the motivation of using language for visual learning,
we focus on learning visual representations by using linguistic guidance from pre-trained language models.

\nsparagraph{Leveraging structure in the data.}
This is commonly done in dense feature learning, where optical flow~\cite{han2020coclr,wang2019learning,jabri2020space,shan2021cohesiv} or 3D transformations~\cite{hou2021pri3d,Wu2021Towers,shang2022learning,elbanani2021unsupervisedr,sun2021loftr} provide natural associations between image patches. 
For images, prior approaches used class names~\cite{roth2022integrating,khosla2020supervised}, class hierarchies~\cite{li2019large,yan2015hd}, meta data~\cite{johnson2015love,li2017ngrams,gong2014multi} or clustering~\cite{caron2018deepcluster,caron2020unsupervised,asano2019sela,tian2021divide,zheng2021weakly} to improve learning and inference. 
Within contrastive learning, clustering has been a popular choice for leveraging dataset structure.
The intuition is that natural clusters emerge in feature spaces that can provide an additional training signal or useful pseudo-labels. 
While such approaches work well on curated datasets (\eg, ImageNet) where the label set provides an estimate of the number of clusters, it struggles with imbalanced and uncurated data~\cite{assran2023hiddenprior}. 
Other approaches sample nearest neighbors as a feature-driven within-domain augmentation~\cite{dwibedi2021nnclr,li2022declip}.
While these approaches differ in how they extract inter-instance relationships, they all use within-domain feature similarity to sample positive pairs or clusters and hence do not leverage the rich cross-modal relationships. 
Closest to our work is Han~\etal~\cite{han2020coclr} who propose a co-training~\cite{blum1998combining} scheme for jointly learning image and optical flow representations. 
We share their motivation of using similarity in one space (language) to learn in another (vision).
Furthermore, instead of relying on co-training on the same dataset, we extract distances from a text-only language model, allowing us to leverage unaligned data.

\section{Method}
\label{sec:method}
The goal of this work is to learn visual representations that can generalize to other datasets.
We extend image-only contrastive learning beyond hand-crafted augmentations and visually-sampled clusters to learn from conceptually similar images.
Through learning to associate images that depict the same \textit{visual concept}, models can learn visual invariances that more closely capture human semantics.
To achieve this, we propose sampling image pairs that have similar captions using a pre-trained sentence encoder~\cite{reimers2019sbert} and using them for contrastive learning.
This work does not propose a new model or loss but rather a novel way of sampling image views that is applicable to a variety of approaches and losses for learning visual representations.

\subsection{Learning from Conceptual Similarity} 

Instance discrimination has been the dominant task for visual representation learning. 
Its core intuition is that visual similarity is a good proxy for semantic similarity.
The standard approach generates positive \textit{view} pairs using image augmentations and maximizes their embedding similarity, with or without negative views. 
While there has been a large number of contrastive learning approaches, view pair generation has largely remained the same. 
Other methods use visual feature similarity to learn prototypes~\cite{asano2019sela,caron2018deepcluster,caron2020unsupervised} or sample previously seen instances~\cite{dwibedi2021nnclr} for contrastive learning. 
While these approaches extend beyond instances and consider relations in the dataset, they still rely on visual similarity to generate their contrastive pairs. 
This limits the visual invariances that they can learn~\cite{xiao2020should}.

We propose training models to identify the same visual \emph{concept} instead of the same \emph{instance}.
Our key observation is simple: images that have similar captions often depict similar concepts regardless of the actual appearance similarity.
This can be clearly seen in~\cref{fig:sampled_pairs}. 
Nearest neighbors in visual representation space depict objects in similar scenes and poses,
with self-supervised models showing some color invariances due to color augmentation.
Conversely, similarly captioned images depict objects in different colors, poses, and contexts. 
This makes language-sampled images an excellent source for visual representation learning as they implicitly capture human-like visual invariances.

\begin{figure*}[ht!]
  \centering
    \includegraphics[width=\linewidth]{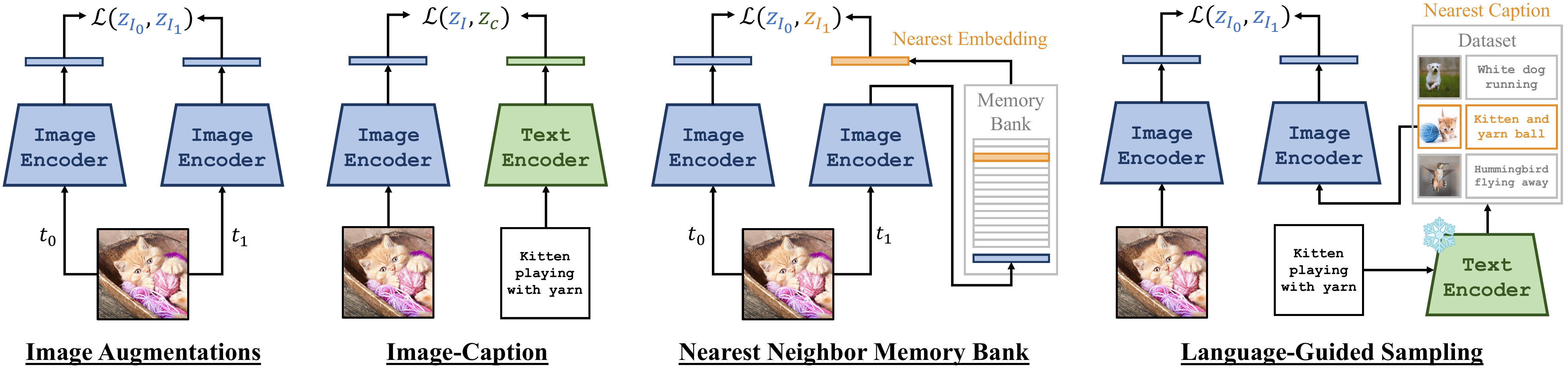}
    \caption{\textbf{Contrasting Contrastive Formulations}. While image-only and image-text contrastive learning directly extract views from the instance, nearest-neighbor methods rely on a memory bank of previously extracted features for training. In contrast, our approach samples nearest neighbors in caption embedding space using a pretrained language model and use the associated image for contrastive learning. }
  \label{fig:method}
\end{figure*}

\subsection{Sampling Image Pairs using Language}

Given a captioned image dataset, we want to sample image pairs that have very similar captions. 
While caption similarity may be a good proxy for conceptual similarity, measuring caption similarity is a challenge on its own. 
Traditional metrics such as BLEU~\cite{papineni2002bleu} and CIDER~\cite{vedantam2015cider} rely on n-gram overlap, which can be too sensitive to phrasing and sentence structure. 
This makes them ill-suited for our needs.
Other metrics such as SPICE~\cite{anderson2016spice} account for such variety by comparing parse trees; however, they still can not account for different wording choices. 
Inspired by advances in language models as well as approaches like BERTScore~\cite{zhang2019bertscore} and CLIPScore~\cite{hessel2021clipscore}, we use a pre-trained sentence encoder to compute caption similarity.

Sentence encoders are trained to extract sentence-level features~\cite{reimers2019sbert,logeswaran2018efficient,kiros2015skip}.
We use SBERT~\cite{reimers2019sbert}, which fine-tunes a pre-trained language model to allow it to better capture semantic similarity using feature cosine distance.
SBERT is trained in two stages: first, a language backbone is trained using a standard self-supervised task such as masked~\cite{mikolov2013distributed,devlin2018bert} or permuted~\cite{song2020mpnet} language modeling;
second, the language modeled is fine-tuned via contrastive learning on a large combined dataset of 1 billion sentence pairs. 
Fine-tuning the model in a contrastive way simplifies downstream usage as it allows features to be compared directly using cosine similarity. 
We use an SBERT~\cite{reimers2019sbert} model with an MPNet~\cite{song2020mpnet} backbone. However, we find that our formulation is not sensitive to the choice of language encoder, as shown in~\cref{tab:sampling_choice}.

Finally, we sample the nearest neighbors for all captions in the language embedding space. 
We leverage modern similarity search libraries~\cite{johnson2019faiss} to perform the nearest neighbor search quickly, despite the large dataset size. 
For example, nearest neighbor sampling runs in under 3 hours for RedCaps (12 million instances) on 4 GPUs, with 43 minutes spent on feature extraction and 117 minutes on nearest neighbor search. 
Furthermore, we find that we could further reduce the complexity of the sampling by only searching within subsets of the data as shown in~\Cref{app:redcaps_shards}.

\subsection{Language-Guided Visual Learning}

Our approach is applicable to several representation learning methods as it only changes the training view pairs.
We focus on contrastive learning since its fairly minimal setting allows us to analyze the impact of language guidance with minimal confounding factors.
We train SimCLR with the language-sampled pairs and refer to it as LGSimCLR.
We also evaluate the impact of language guidance on SimSiam~\cite{chen2021exploring} and SLIP~\cite{mu2021slip}, and find that they can similarly benefit from language guidance.
We only use random cropping for image augmentations since language-sampled pairs are naturally augmented versions of each other and find that additional augmentations are not helpful. 
For LGSLIP, we match their setup by applying the CLIP loss only between the source's image and caption, ignoring an additional loss between the nearest neighbor image and its caption.

\section{Experiments}
\label{sec:experiments}
Our experiments evaluate the efficacy of learning visual features from conceptually similar images.
We hypothesize that a model trained with language guidance will learn useful visual invariances and better generalize to downstream tasks.
We are interested in answering these questions: 
Does language guidance improve generalization over other pre-training approaches? 
Does language guidance generalize to other datasets and pre-training approaches?
How can language be used for visual pre-training?

\subsection{Experimental Setup}

We formulate our experimental setup to compare the efficacy of different learning signals.
We train models with language-guided sampling and compare them with image-only self-supervised models and image-text contrastive models.
We are interested in conducting controlled experiments for a fair comparison.

Recent work in self-supervised learning has demonstrated the impressive impact of scaling~\cite{radford2021learning,chen2020simple,zhai2022scaling}. 
While such work has shown impressive performance, it has complicated the evaluation as different models are trained on different pretext tasks on different datasets using varying amounts of compute and training recipes. 
Furthermore, replication is difficult, if not impossible, due to the unavailability of training data or prohibitive compute requirements. 
Fortunately, several papers report results that indicate that performance patterns often hold at smaller scales~\cite{radford2021learning,caron2020unsupervised,chen2020simple,mu2021slip,cui2022democratizing}. 
Hence, we conduct our experiments at a scale that allows us to perform a comprehensive evaluation and permits replication by others.

We conduct our experiments with a standard backbone~\cite{he2016deep} on publicly available datasets~\cite{cc3m,cc12m,desai2021redcaps}. 
To account for variation in training recipes, we retrain all methods from scratch using the same training recipe. 
We scale down experiments to a level that permits fair comparisons and replication.
We also provide system-level comparisons in~\cref{tab:apples_to_oranges} and scaling results in~\cref{app:scaling}.

% Training Details
\llsparagraph{Training details:}
We use a ResNet-50 backbone and train all models using the AdamW optimizer~\cite{loshchilov2019decoupled} with a learning rate of $10^{-3}$ and a weight decay of $10^{-2}$. 
We use a cosine learning scheduler~\cite{loshchilov2016sgdr} with 5000 warm-up steps. 
Models are trained using a batch size of 512 for 250k steps; this corresponds to 10.5 epochs on RedCaps.
We use a constant number of steps to permit meaningful comparisons between models trained on different datasets.

% Evaluation 
\llsparagraph{Evaluation setup:}
We evaluate all approaches using linear probe and fewshot classification on 15 classification datasets inspired by~\cite{kornblith2019better,radford2021learning}. 
We use the linear probe evaluation proposed by~\cite{kornblith2019better} and learn a single linear layer using logistic regression. We sweep over a range of cost values and choose the value with the best validation performance. 
We retrain a classifier on both train and validation splits and report test performance. 
We also evaluate all approaches on fewshot classification to understand their generalization ability. 
We use a weighted kNN classifier on frozen support features inspired by prior work showing its effectiveness for fewshot classification~\cite{wang2019simpleshot}.
Please see~\Cref{app:evaluation_tasks,app:evaluation_datasets} for more details on evaluation datasets and tasks.

\lsparagraph{Baselines:} 
While there have been many proposed visual representation learning approaches, they can be grouped into several key directions that differ in the pretext task.
We focus our comparison on a few representative approaches to explore the impact of the learning signal.
We overview the baselines here and provide more details in~\Cref{app:baselines}.

Many of our baselines are variants of contrastive learning as shown in~\cref{fig:method}.
Contrastive approaches operate over paired source and target feature embeddings: $\bm z^{s}$ and  $\bm z^{t}$.
The goal is to maximize the similarity between the paired embeddings and minimize it with respect to all other embeddings. 
Given a batch size $N$ and embedding dimension $F$, $\bm z^{s}, \bm z^{t} \in \mathbb{R}^{N \times F}$.  The contrastive loss~\cite{sohn2016improved} is: 
\begin{equation}
    \mathcal{L}(z^{s}, z^{t}) = 
    -\log \frac{\exp(\mathrm{sim}(\bm z^s_i, \bm z^t_i)/\tau)}{\sum_{k=1}^{N} \exp(\mathrm{sim}(\bm z^s_i, \bm z^t_k)/\tau)}~,
\end{equation}
where $\tau$ is a scaling parameter and $\mathrm{sim}(\cdot,\cdot)$ is cosine similarity. 
Contrastive approaches primarily differ in how the embeddings are computed.

\llsparagraph{Image-Only Contrastive Learning} contrasts features extracted from two randomly augmented \textit{views} of the same image to perform instance discrimination~\cite{wu2018unsupervised}. 
We use SimCLR~\cite{chen2020simple} as a representative approach due to its simplicity and strong performance.

\llsparagraph{Image-Text Contrastive Learning} learns by contrasting features extracted from images and their captions. 
Unlike image-only approaches, this approach can learn semantics from the captions. 
Radford~\etal~\cite{radford2021learning} first proposed this approach and has had several follow-ups that augment it with additional self-supervised losses losses~\cite{mu2021slip,li2022declip,lee2022uniclip}. 
We use CLIP~\cite{radford2021learning} and SLIP~\cite{mu2021slip} due to their simplicity.

\lsparagraph{Nearest Neighbor Contrastive Learning} contrast source embeddings with retrieved embeddings from a memory bank. 
The target features are used to retrieve the nearest neighbor embedding from a memory bank of previous batches.
Dwibedi~\etal~\cite{dwibedi2021nnclr} proposed this approach for image-only contrastive learning, while Li~\etal~\cite{li2022declip} proposed adapting this loss for language embeddings. 
We use NNCLR~\cite{dwibedi2021nnclr} as Visual NNCLR and DeCLIP~\cite{li2022declip} with the CLIP and the language NNS losses as Language NNCLR.

\lsparagraph{Image-Only Non-Contrastive Learning} deviates from the typical contrastive setup by learning without negative samples~\cite{grill2020bootstrap,chen2021exploring}.  
We use SimSiam as a representative approach due to its simplicity and strong performance. 

\lsparagraph{Cluster-based Contrastive Learning} learn by contrasting image features with learned prototypes~\cite{caron2018deepcluster,caron2020unsupervised,asano2019sela}. 
Prototypes are estimated via clustering or learned jointly with the feature encoder. 
Caron~\etal~\cite{caron2020unsupervised} report that different cluster-based approaches perform similarly when provided with the same algorithmic  advances. 
We use an adapted SwAV without the multi-crop augmentation strategy as it is equally applicable to other methods. 
We also compare against a pre-trained SwAV checkpoint in~\cref{tab:apples_to_oranges}.

\begin{table*}
  \centering
    \caption{
  \textbf{Linear Probe Evaluations.} We train ResNet-50 models on RedCaps and report performance of a linear probe using frozen features on 15 downstream tasks. Models are split based on whether or not they require caption images for training. LGSimCLR outperforms all previous approaches with strong performance gains for fine-grained classification datasets. }
  \label{tab:linear_probe}
  \setlength\tabcolsep{4pt}
  \small
  \begin{tabularx}{\linewidth}{X cccccccccccccccc}
    \toprule
    \textbf{Model} &
    \rotatebox[origin=lb]{90}{\bf {Food-101}} &
    \rotatebox[origin=lb]{90}{\bf {CIFAR-10}} &
    \rotatebox[origin=lb]{90}{\bf {CIFAR-100}} &
    \rotatebox[origin=lb]{90}{\bf {CUB}} &
    \rotatebox[origin=lb]{90}{\bf {SUN397}} &
    \rotatebox[origin=lb]{90}{\bf {Cars}} &
    \rotatebox[origin=lb]{90}{\bf {Aircraft}} &
    \rotatebox[origin=lb]{90}{\bf {DTD}} &
    \rotatebox[origin=lb]{90}{\bf {Pets}} &
    \rotatebox[origin=lb]{90}{\bf {Caltech-101~}} &
    \rotatebox[origin=lb]{90}{\bf {Flowers}} &
    \rotatebox[origin=lb]{90}{\bf {STL-10}} &
    \rotatebox[origin=lb]{90}{\bf {EuroSAT}} &
    \rotatebox[origin=lb]{90}{\bf {RESISC45 }} &
    \rotatebox[origin=lb]{90}{\bf {PCAM}} &
    \rotatebox[origin=lb]{0}{\bf {Mean}} \\
    \midrule
SwAV                 &  63.6 &  81.3 &  57.5 &  21.6 &  47.5 &  22.9 &  35.4 &  68.1 &  61.1 &  70.5 &  78.0 &  87.7 &  94.3 &  79.9 &  84.3 & 63.6 \\
SimSiam              &  64.1 &  79.9 &  56.1 &  28.2 &  48.3 &  29.5 &  41.2 &  66.2 &  69.1 &  73.6 &  83.6 &  85.7 &  94.4 &  82.1 &  83.3 & 65.7 \\
Visual NNCLR         &  65.4 &  82.8 &  60.2 &  26.6 &  50.0 &  26.6 &  40.9 &  68.0 &  65.2 &  75.4 &  83.5 &  88.5 &  95.3 &  82.2 &  83.8 & 66.3 \\
SimCLR               &  69.0 &  82.9 &  61.6 &  30.6 &  52.6 &  33.7 &  43.7 &  69.8 &  70.5 &  74.1 &  86.9 &  88.0 &  95.4 &  84.6 &  84.4 & 68.5 \\
\midrule
Language NNCLR       &  81.2 &  83.1 &  61.9 &  48.6 &  56.5 &  45.1 &  37.2 &  68.8 &  78.1 &  82.0 &  90.2 &  93.4 &  92.5 &  81.1 &  80.7 & 72.0 \\
CLIP                 &  80.9 &  84.7 &  62.7 &  50.4 &  57.4 &  45.8 &  36.7 &  67.6 &  79.8 &  84.0 &  91.0 &  93.5 &  93.9 &  82.2 &  82.6 & 72.9 \\
SLIP                 &  77.7 &  87.2 &  67.0 &  42.4 &  58.1 &  48.7 &  45.2 &  72.3 &  79.5 &  82.7 &  92.1 &  92.7 &  95.6 &  85.5 &  83.4 & 74.0 \\
LGSimCLR (Ours)      &  83.2 &  87.8 &  69.0 &  59.3 &  60.3 &  62.3 &  53.4 &  71.2 &  81.8 &  89.4 &  95.9 &  94.0 &  95.6 &  88.0 &  81.1 & 78.2 \\
    \bottomrule
  \end{tabularx}
\end{table*}

% ======================== Linear Probe Classification ===========================
\subsection{Results}

We train all approaches with a ResNet-50 backbone on RedCaps and report results in~\cref{tab:linear_probe,tab:fewshot}.
Our model outperforms all baselines with a significant margin for both evaluations. We analyze the results below through a series of questions.

\lsparagraph{Does language-guided sampling provide better training pairs than image augmentations? }
LGSimCLR greatly outperforms SimCLR despite using the same learning objective. 
By using language sampled pairs instead of image augmentations, LGSimCLR learns stronger invariances. 
We find that the largest gains arise in fine-grained datasets: Cars, CUB, and Food101. 
The performance gains can be explained by considering the critique of Xiao~\etal~\cite{xiao2020should}: the training augmentations dictate the invariances learned by SimCLR as shown in nearest neighbor samples in~\cref{fig:sampled_pairs}.
Consider the third row of~\cref{fig:sampled_pairs}, while language sampling depicts three Aston Martin cars in different spots, visual nearest neighbors are sports cars in different poses and colors, closely resembling the flip and color augmentations used for training.  
Similarly in the first row of~\cref{fig:sampled_pairs}, visual nearest neighbors depict owls from different species in similar poses, while language sampling retrieves three great horned owls from different viewpoints. 
These trends are further amplified when features are used directly for fewshot classification. 
Language guidance allows us to capture relationships that go beyond visual similarity by training on image pairs that capture human semantics. 

\begin{figure}[t!]
  \centering
  \includegraphics[width=0.95\linewidth]{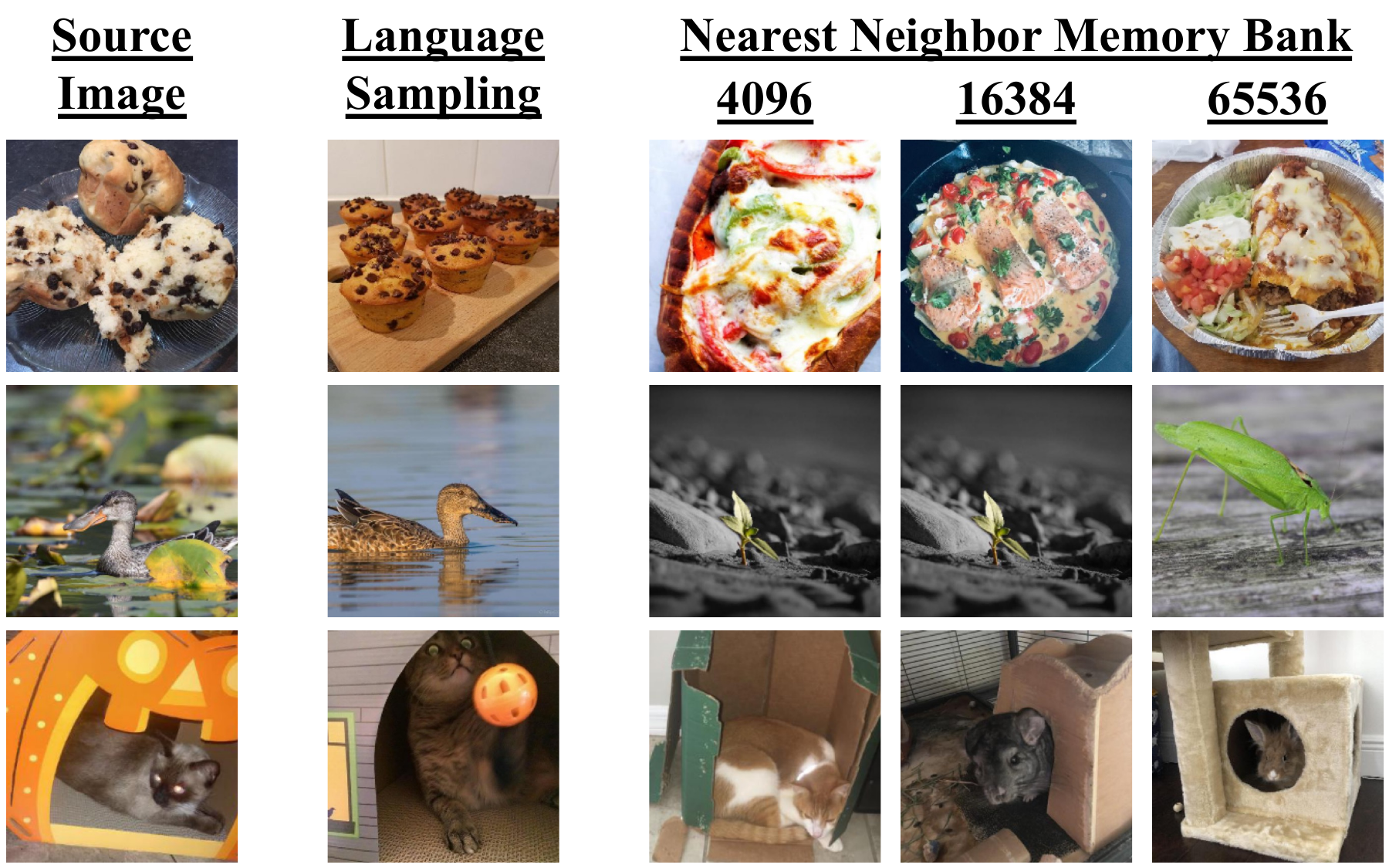}
  \caption{\textbf{Nearest Neighbor methods are limited by the memory bank size.} 
  Even with a large memory bank, the nearest embedding can still be unrelated to the source image while language sampling provides us with conceptually similar pairs. \vspace{-1em}}
  \label{fig:language_vs_queue}
\end{figure}

\lsparagraph{Can we just sample nearest neighbors from previous batches?}
LGSimCLR outperforms NNCLR despite both relying on nearest neighbors. 
NNCLR uses the nearest feature embedding from a memory bank in the same modality. 
The quality of their retrieved samples is limited by the size of the memory bank. 
To demonstrate this, we visualize the nearest neighbors retrieved by NNCLR for different memory bank sizes in ~\cref{fig:language_vs_queue}.
We find that the retrieval quality is poor even for larger queues. 
Interestingly, we note that NNCLR also underperforms SimCLR on RedCaps, despite performing better on ImageNet. 
We posit that ImageNet's curated distribution explains this: a queue of 16k will most probably contain instances from each class, resulting in both visually and conceptually similar retrievals.
Additionally, the quality of nearest neighbors is affected by the sampling feature space; features that are only trained on image augmentations will have limited invariances as shown in~\cref{fig:sampled_pairs}.
We further explore the impact of sampling space on training in ~\cref{sec:analysis}.

\begin{table}
  \caption{\textbf{Language-guided contrastive learning outperforms image-text contrastive learning, regardless of text encoder.}}
  \label{tab:clip_sbert}
  \centering
  \small
  \setlength\tabcolsep{4pt}
  \begin{tabularx}{\linewidth}{X l cc}
    \toprule
    Objective           & Text Encoder      & Linear     & Fewshot \\
    \midrule 
    \multirow{2}{*}{Image-Text}                        
            & Randomly-Initialized          & 72.9 & 77.5 \\
            & Frozen SBERT                  & 71.8 & 77.1 \\
    \midrule
    \multirow{2}{*}{Image-Image}             
            & Frozen CLIP (RedCaps)       & 78.3 & 82.4 \\
            & Frozen SBERT         & 78.2 & 82.5 \\
    \bottomrule
  \end{tabularx}
\end{table}

\begin{table*}
  \caption{\textbf{Few-Shot Evaluations.} We train ResNet-50 models on RedCaps and report 5-way, 5-shot classification performance.
  We observe that language results in huge performance gains as shown by the performance of CLIP and LGSimCLR. 
  Furthermore, the use of any augmentations hurts performance as seen by SLIP's drop in performance. 
  }

  \label{tab:fewshot}
  \centering
  \setlength\tabcolsep{4pt}
  \small
  \begin{tabularx}{\linewidth}{X cccccccccccccccc}
    \toprule
    \textbf{Model} &
    \rotatebox[origin=lb]{90}{\bf {Food-101}} &
    \rotatebox[origin=lb]{90}{\bf {CIFAR-10}} &
    \rotatebox[origin=lb]{90}{\bf {CIFAR-100}} &
    \rotatebox[origin=lb]{90}{\bf {CUB}} &
    \rotatebox[origin=lb]{90}{\bf {SUN397}} &
    \rotatebox[origin=lb]{90}{\bf {Cars}} &
    \rotatebox[origin=lb]{90}{\bf {Aircraft}} &
    \rotatebox[origin=lb]{90}{\bf {DTD}} &
    \rotatebox[origin=lb]{90}{\bf {Pets}} &
    \rotatebox[origin=lb]{90}{\bf {Caltech-101~}} &
    \rotatebox[origin=lb]{90}{\bf {Flowers}} &
    \rotatebox[origin=lb]{90}{\bf {STL-10}} &
    \rotatebox[origin=lb]{90}{\bf {EuroSAT}} &
    \rotatebox[origin=lb]{90}{\bf {RESISC45 }} &
    \textbf{Mean}         \\
    \midrule

SwAV                 & 64.5            & 54.0            & 61.8            & 45.8            & 84.9            & 36.5            & 34.1            & 74.8            & 66.5            & 78.1            & 75.5            & 72.6            & 80.4            & 72.9            & 64.5 \\
SimSiam              & 63.9            & 49.9            & 57.2            & 49.5            & 84.5            & 39.3            & 37.9            & 75.7            & 67.8            & 79.7            & 81.5            & 69.6            & 80.6            & 79.4            & 65.5 \\
Visual NNCLR         & 65.6            & 54.1            & 61.7            & 45.8            & 85.3            & 37.9            & 34.9            & 75.2            & 67.3            & 81.1            & 75.4            & 74.3            & 83.6            & 76.7            & 65.6 \\
SimCLR               & 66.9            & 45.7            & 51.0            & 51.5            & 87.1            & 44.0            & 38.4            & 77.6            & 70.1            & 80.0            & 86.9            & 69.6            & 83.5            & 81.3            & 66.7 \\
\midrule
Language NNCLR       & 89.3            & 65.3            & 73.4            & 78.6            & 90.8            & 68.4            & 40.4            & 75.2            & 78.8            & 90.9            & 94.3            & 89.6            & 75.2            & 71.9            & 77.3 \\
CLIP                 & 88.9            & 64.6            & 73.1            & 78.3            & 90.9            & 69.7            & 40.7            & 75.7            & 77.5            & 91.6            & 94.7            & 89.8            & 75.3            & 74.8            & 77.5 \\
SLIP                 & 81.5            & 63.5            & 70.8            & 63.1            & 91.3            & 62.9            & 42.1            & 79.6            & 76.4            & 88.4            & 92.2            & 83.4            & 82.7            & 80.8            & 75.6 \\
LGSimCLR (Ours)      & 90.3            & 66.3            & 75.5            & 83.1            & 92.7            & 77.6            & 50.6            & 81.1            & 84.1            & 95.4            & 97.6            & 86.5            & 85.0            & 89.0            & 82.5 \\
    \bottomrule
  \end{tabularx}
 % }
\end{table*}

\llsparagraph{Can cluster-based approaches learn better features?} 
Similar to nearest-neighbor sampling, clustering is performed using visual similarity. 
Furthermore, it is based on an estimated number of clusters in the training dataset. 
Although this can be determined for ImageNet due to its known class structure, the number of clusters in an arbitrary uncurated dataset is unknown. 
This results in a large performance drop, as seen in~\cref{tab:linear_probe} and~\cref{tab:fewshot}.
On the other hand, sampling related pairs assumes no global structure within the data and hence is able to better capture inter-instance similarity. 
This results in nearest-neighbor sampling outperforming clustering and both being outperformed by contrastive learning and language-guided contrastive learning.

\lsparagraph{Should we use language for guidance or supervision?}
Our experiments indicate that LGSimCLR outperforms both CLIP and SLIP. 
We consider two possible explanations:
(a) SBERT extracts better language embeddings than CLIP can learn from the data, or
(b) language-guided contrastive learning is a better training objective than image-text contrastive learning. 
To evaluate this, we compare four models in~\cref{tab:clip_sbert}. 
The first two models use CLIP's training objective: the first model uses a randomly initialized language encoder, similar to CLIP. 
The second model uses a frozen SBERT model as the language encoder and only trains the projection layers. 
The second two models use LGSimCLR's training objective but sample pairs using a pre-trained language-only SBERT or the language encoder from a CLIP model trained on RedCaps. 
We find that image-image contrastive learning yields better visual features for both setups. 
While CLIP does not benefit from an SBERT backbone, LGSimCLR benefits from sampling using a language encoder trained on the same dataset.
This suggests that learning joint embeddings results in worse visual features than language-guided learning.

\lsparagraph{System-level comparisons: } 
We compare LGSimCLR with publicly-available checkpoints of prior approaches; see~\Cref{app:baselines} for details.
We emphasize that while the experiments reported in~\cref{tab:linear_probe,tab:fewshot} were done in a controlled setup (same batch size, training data, optimizer), the system level comparisons are trained on different datasets with different training recipes and enhancements to further boost performance; \eg, large batch sizes, longer training, multi-crop augmentation.
Furthermore, it has been shown that models trained on ImageNet implicitly benefit from its curated nature~\cite{mu2021slip,assran2023hiddenprior}. 
Nevertheless, our approach still outperforms prior self-supervised approaches. 
We fall short of CLIP's ResNet-50 due to its training scale; $64{\times}$ larger batch, $32{\times}$ larger dataset, and 75${\times}$.
We also observe that ImageNet-supervised ResNet-50 achieves better fewshot performance. 
Examining the performance breakdown in~\cref{tab:app_fewshot}, we find the improvement mainly comes from CIFAR10, CIFAR100, and Pets. We posit that this can be explained by ImageNet's class structure: mostly pets with a large overlap with CIFAR's classes. 

\begin{table}
  \caption{
  \textbf{ResNet-50 System Level Comparisons.} We outperform prior self-supervised approaches despite them benefiting from ImageNet's curation for training and using larger batch sizes. CLIP outperforms us due to the scale of its training. 
  }
  \label{tab:apples_to_oranges}
  \centering
  \setlength\tabcolsep{3pt}
\footnotesize
\begin{tabularx}{\linewidth}{X ccccc}

\toprule
                    & Batch      & \# Img Updates              & Dataset     & Linear            & Fewshot  \\
\midrule 

Supervised~\cite{wightman2021resnet}               & 1024  & $1.3{\times}10^8$    & ImageNet & 78.0 & 85.7 \\
SimSiam~\cite{chen2021exploring}                   & 512   & $1.3{\times}10^8$    & ImageNet & 72.9 & 78.7 \\
SimCLR~\cite{chen2020big}                          & 4096  & $1.0{\times}10^9$    & ImageNet & 75.4 & 77.4 \\
MoCo~\cite{chen2021mocov3}                         & 4096  & $1.3{\times}10^8$    & ImageNet & 77.7 & 80.1 \\
SwAV~\cite{caron2020unsupervised}                  & 4096  & $1.3{\times}10^8$    & ImageNet & 78.2 & 78.5 \\
CLIP~\cite{radford2021learning}                    & 32768 & $1.0{\times}10^{10}$   & CLIP     & 81.8 & 87.8 \\
LGSimCLR                                           & 512   & $1.3{\times}10^8$    & RedCaps  & 78.2 & 82.5 \\
\bottomrule
\end{tabularx}

\end{table}

\begin{table*}
\subcaptionbox{\textbf{Approach Generality} \label{tab:approach_generality}}{
  \centering
  \footnotesize
  \setlength\tabcolsep{2.5pt}
  \begin{tabularx}{0.33\linewidth}{X cc cc cc}
    \toprule
    & \multicolumn{2}{c}{Image Aug.}  
    & \multicolumn{2}{c}{Language} 
    \\
    \cmidrule(lr){2-3} 
    \cmidrule(lr){4-5}
                    & Linear     & Fewshot    & Linear     & Fewshot      \\
    \midrule 
    SimSiam              & 65.7 & 65.5 & 71.2 & 75.7 \\
    SimCLR               & 68.5 & 66.7 & 78.2 & 82.5 \\
    SLIP                 & 74.0 & 75.6 & 78.8 & 82.8 \\
    \bottomrule
  \end{tabularx}
  % }
}
\hfill
\subcaptionbox{\textbf{Impact of training dataset} \label{tab:dataset_impact}}{
  \centering
  \footnotesize
  \setlength\tabcolsep{2.5pt}
  \begin{tabularx}{0.3\linewidth}{X cc cc}
    \toprule
                        & Size      & Linear    & Fewshot \\
    \midrule 
    CC3M           &  2.7M & 71.5 & 76.3 \\
    CC12M          & 10.9M & 76.8 & 81.9 \\
    RedCaps 2020   &  3.2M & 73.8 & 78.8 \\
    RedCaps        & 12.0M & 78.2 & 82.5 \\

    \bottomrule
  \end{tabularx}
  }
  % }
\hfill
\subcaptionbox{\textbf{Impact of sampling space}  \label{tab:sampling_choice}}{
  \centering
  \footnotesize
  \setlength\tabcolsep{2.5pt}
  \begin{tabularx}{0.31\linewidth}{X cc}
    \toprule
                             & Linear & Fewshot \\
    \midrule 
    SBERT (MPNet)            & 78.2 & 82.5 \\
    SBERT (MiniLM)           & 78.6 & 83.3 \\
    CLIP Language (ViT-B/32) & 78.3 & 83.1 \\
    FastText BoW             & 76.1 & 80.9 \\
    \midrule
    ImageNet-supervised      & 78.3 & 81.8 \\
    SimCLR (ImageNet)        & 73.1 & 74.6 \\
    \bottomrule
  \end{tabularx}
  }
% }

\vspace{-2pt}
\caption{\textbf{Analysis Experiments.} We conduct a series of analysis experiments to understand language-guided contrastive learning. The results indicate that language sampling is beneficial to several formulations and scales well with larger datasets. Furthermore, while language sampling consistently results in good pairs for training, visual sampling only helps if it has access to semantics through supervision. }
\vspace{-6pt}
\end{table*}

% ======================== Sampling Choice Analysis ===========================
\subsection{Analysis}
\label{sec:analysis}

We now analyze language-guided contrastive learning by evaluating the impact of pre-training data, the choice of embedding space, and the pretext task. By understanding the impact of those choices, we can better understand what the model is learning. 

\lsparagraph{Approach generality:}
We extend language guidance to other contrastive approaches: SimSiam and SLIP. 
We observe that language guidance uniformly improves performance for all methods, as shown in~\cref{tab:approach_generality}. 
Furthermore, the difference between SimCLR and SLIP shrinks when adding language guidance. 
This suggests that language guidance provides the model with similar semantics to the ones learned from an image-text contrastive loss, resulting in diminished gains from the additional image-text loss.

\lsparagraph{Impact of training dataset:}
We train our model on four datasets:
CC3M~\cite{cc3m}, CC12M~\cite{cc12m}, RedCaps-2020, and RedCaps~\cite{desai2021redcaps}.
In \cref{tab:dataset_impact}, we observe that larger datasets result in stronger performance, indicating that our approach could scale well with even larger datasets.
Furthermore, we observe that RedCaps results in better performance than Conceptual Captions. 
This may be attributed to the higher quality of captions in RedCaps;
while the alt-text captions CC3M and CC12M can be short and contain image metadata, RedCaps captions are diverse, longer, and more descriptive.
This allows our model to sample more interesting visual pairs that capture more visual diversity.
We provide qualitative results in~\Cref{app:qualitative} to support this. 

% \vspace{0.1cm}
\llsparagraph{Impact of sampling space:}
The idea of using offline nearest-neighbor sampling does not require a specific language model or even a specific modality. 
We explore other choices for embedding space: four sentence encoders and two image models. 
In our experiments, we use SBERT's MPNet model~\cite{reimers2019sbert,song2020mpnet}; the highest performing SBERT model for sentence similarity.
We compare it to two other sentence transformers:
a smaller SBERT model, MiniLM~\cite{wang2020minilm}, and the language encoder from CLIP~\cite{radford2021learning}.
We also compared against a bag-of-words (BoW) sentence encoder that uses FastText~\cite{bojanowski2016enriching} embeddings.
Results are in \cref{tab:sampling_choice}.
While we expected that using CLIP for sampling would improve performance due to its multimodal training, we were surprised that MiniLM also improved performance despite its lower performance on language tasks.
We find that pairs obtained using a BoW model result in a weaker performance which might hint at the importance of contextual sentence embeddings.
Nevertheless, the BoW-sampled pairs still result in higher performance than all the other baselines on RedCaps. 

We also consider training with pairs sampled using two visual models: ImageNet-supervised ResNet-50~\cite{wightman2021resnet} and ImageNet-trained SimCLR~\cite{chen2020big}.
We find that using a visual model for sampling is only beneficial if the visual model captures semantic relations; \eg, through supervised training.
Using a self-supervised language model results in a strong drop in performance relative to the other sampling spaces. 
Nevertheless, it still allows the model to achieve better performance than using a self-supervised visual approach on the same data. This indicates that while language is a better modality to use, ``sample-guided'' contrastive learning can still achieve a stronger performance than only using self-supervised learning.

\llsparagraph{Limitations:}
We observe a few limitations in our approach. 
Image captions can be noisy, vague, and often omit obvious relations in the image~\cite{bagherinezhad2016elephants}.
While this broadly affects image-language models, it can result in us retrieving unrelated image pairs.
For example, captions like ``{\textit{I found this in the garden}}'' or~``{\textit{Photo from our family trip}}'' could describe a large range of images, some of which are unrelated. 
We expand on this in~\Cref{app:qualitative}.
Image descriptions also depend on the context and the perceiver; \eg, a tourist and an art curator will describe artwork in very different ways.
We observe that descriptions in topic-focused subreddits (\eg, \textbf{\texttt{r/birdpics}} and \textbf{\texttt{r/woodworking}}) are more specific than in generic subreddits (\eg,\texttt{\textbf{r/itookapicture}} and \texttt{\textbf{r/pics}}).
Our experiments in~\Cref{app:redcaps_shards} support this observation.
Since a caption only captures one aspect of the image, sampled pairs can be similar for a variety of reasons. 
Allowing the model to condition the feature extraction or similarity calculation on captions could alleviate this issue.

\section{Conclusion}
\label{sec:conclusion}
We propose using language to find conceptually similar images for contrastive learning.
This is based on a simple observation: people describe an object in similar ways even when it appears in different contexts. 
We use pre-trained language models to sample similar captions and use the captioned images for contrastive learning. 
We hypothesize that using language guidance instead of image augmentations would result in learning more human-like invariances.

We evaluate our approach on multiple train and test datasets and find that it outperforms previous self-supervised and image-text contrastive models. 
Our analysis demonstrates the utility of using nearest-neighbor instances for training and the superiority of language sampling over other approaches for unlabeled datasets. 
Our findings align with prior work that critiques the use of image augmentations~\cite{xiao2020should,selvaraju2021cast} and shows the utility of cross-modal guidance~\cite{han2020coclr} and intra-instance relationships~\cite{khosla2020supervised,dwibedi2021nnclr}.
Our results demonstrate the potential of incorporating language guidance in contrastive learning.
We hope that future work will explore scaling up our approach to larger and more diverse datasets, as well as modeling approaches that further integrate language into the learning process. 

{
\small
\vspace{0.2cm}
\noindent 
\textbf{Acknowledgments:}
We thank Richard Higgins, Ashkan Kazemi, and Santiago Castro for many helpful discussions, as well as David Fouhey, Ziyang Chen, Chenhao Zheng, Fahad Kamran, and Dandan Shan for their feedback on early drafts. This project was funded under the Ford-UM Alliance partnership; we thank Alireza Rahimpour, Devesh Upadhyay, and Ali Hassani from Ford Research for their support and discussion.
}

{
\small
\bibliographystyle{misc/ieee_fullname}
\bibliography{references}
}

\clearpage
\appendix

\section{Evaluation Tasks}
\label{app:evaluation_tasks}
We compare all models by evaluating the encoder's frozen features on two downstream classification tasks: linear probe and fewshot. 
We chose to use simple classifiers since they allow us to evaluate the features as-is and conduct a comprehensive hyperparameter sweep to ensure a fair comparison. We explain the two evaluation setups in more detail below. Our implementation can be found at \url{https://github.com/mbanani/lgssl}. 

\subsubsection*{Linear Probe Classification}

We follow the linear probe evaluation proposed by Kornblith~\etal~\cite{kornblith2019better} of training a logistic regression classifier using the L-BFGS optimizer~\cite{liu1989limited}.
We follow prior work~\cite{kornblith2019better,radford2021learning} and perform a hyperparameter sweep over the cost values in the logistic regression loss.
We sweep over 96 values in log space from $10^{-6}$ to $10^6$. During the hyperparameter sweep, we train on the train split and validate on the valid split. We choose the cost value with the best validation performance and train a final classifier on the combined train and validation instances. We use the PyTorch~\cite{paszke2019pytorch} implementation of L-BFGS with all the default parameters except for the maximum number of iterations, which is set to 1000 similar to CLIP~\cite{radford2021learning}. Our evaluation metric depends on the dataset, as shown in~\cref{table:eval_datasets}, to account for class imbalance. 

\subsubsection*{Few-Shot Classification}

We also use fewshot classification as an evaluation for frozen features. 
Prior work~\cite{wang2019simpleshot,tian2020rethinking} has shown that simple classifiers on top of frozen features are strong baselines for fewshot classification. 
More specifically, Wang~\etal~\cite{wang2019simpleshot} shows that when features are normalized (mean subtraction and L2 normalization), a nearest neighbor classifier is a very effective and strong baseline for fewshot classification. 
Inspired by these results, we use a simple weighted nearest neighbor classifier to evaluate pre-trained frozen features. 
We set $k$ to be the size of the support set and classify the features as follows: 
\begin{align}
    y' = \argmax_{v} \sum_{(I, y) \in \mathbb{D}_{\text{support}}} \indicatorfn{v = y} \mathrm{sim}(f(I'), f(I))
\end{align}
where $\indicatorfn{v = y}$ is an indicator variable that is $1$ if $y$ is the same class as $v$ and 0 other wise, $\mathrm{sim}(\cdot,\cdot)$ is cosine similarity between two vectors, $f$ is the visual encoder, $I'$ is the target image, $\mathbb{D}_{\text{support}}$ is the support set. 

We adopt 5-way, 5-shot classification as our fewshot classification task. We sample five random classes for each episode and then sample five images for each class in the training set, resulting in 25 labeled training images. We also sample 5 images for each class from the test set as our test images. We use all available test images for classes with less than five test images for that class. This is primarily an issue for Caltech-101~\cite{feifei2004caltech101}. We sample 5000 episodes and compute the average test accuracy across all episodes. We experimented with increasing the number of episodes to 50000 to improve evaluation but noticed little change in the mean performance. We also report the 95\% confidence interval for each dataset. 

\section{Evaluation Datasets}
\label{app:evaluation_datasets}
We list all the evaluation datasets used in~\cref{table:eval_datasets}.  
We use TensorFlow datasets for evaluation to ensure easy replication~\cite{TFDS}. 
For all datasets, we preprocess the images by resizing the image so that its smaller dimension is 224 using bilinear interpolation followed by a center crop to $224{\times}224$.
We use bilinear interpolation since improves performance on low-resolution datasets such as CIFAR-10 and CIFAR-100. 
We normalize the images using ImageNet's mean and standard deviation for pixel values for all models except for pre-trained CLIP. 
For CLIP, we use their provided mean and standard deviation values as they greatly impact performance: an average gain of approximately $4\%$ for linear probe evaluation. 
We exclude Patch Camelyon from the fewshot evaluation since it is a binary classification dataset.
We also include statistics for the ImageNet dataset evaluations done in~\Cref{app:results}.

\begin{table}
    \caption{
        \textbf{Evaluation Datasets. }
        Orange rows indicate datasets that do not have an official validation split; we constructed one by randomly holding out 20\% of the official train split.
        Blue rows indicate datasets that do not officially define splits; we randomly sample instances to construct non-overlapping splits.
        For ImageNet, the validation set is used as the test set, and we construct a validation set by randomly holding out 20\% of the official training split. 
        ImageNet variants, highlighted in green, are all test sets for models trained on ImageNet. 
    }
    \label{table:eval_datasets}

    \newcommand{\tblroworange}{\rowcolor{Orange!15}}
    \newcommand{\tblrowblue}{\rowcolor{RoyalBlue!15}}
    \newcommand{\tblrowgreen}{\rowcolor{Green!15}}
    \setlength{\tabcolsep}{2pt}
    \footnotesize
    \begin{tabularx}{\linewidth}{X ccccc}
    \toprule
    \textbf{Dataset} & \textbf{Classes} & \textbf{Train} & \textbf{Val} & \textbf{Test} & \textbf{Metric} \\
    \midrule
    \tblroworange Food-101~\cite{bossard2014food101}        & 101   & 60600 & 15150 & 25250 & accuracy \\
    \tblroworange CIFAR-10~\cite{krizhevsky2009cifar}       & 10    & 40000 & 10000 & 10000 & accuracy \\
    \tblroworange CIFAR-100~\cite{krizhevsky2009cifar}      & 100   & 40000 & 10000 & 10000 & accuracy \\
    \tblroworange CUB-2011~\cite{cub200}                    & 200   & 5795  & 1199  & 5794  & accuracy \\
    \tblroworange SUN397~\cite{xiao2010sun397}              & 397   & 15880 & 3970  & 19849 & accuracy \\
    \tblroworange Stanford Cars~\cite{krause2013cars196}    & 196   & 6515  & 1629  & 8041 & accuracy \\
    FGVC Aircraft~\cite{maji2013aircrafts}                  & 100   & 3334  & 3333  & 3333 & mean-per-cls \\
    DTD~\cite{cimpoi14describing}    & 47    & 1880  & 1880  & 1880 & accuracy \\
    \tblroworange Oxford-IIIT Pets~\cite{parkhi2012pet}     & 37    & 2944  & 736   & 3669 & mean-per-cls \\
    \tblroworange Caltech-101~\cite{feifei2004caltech101}   & 102   & 2448  & 612   & 6084 & mean-per-cls \\
    Oxford Flowers~\cite{nilsback2008flowers}           & 102   & 1020  & 1020 & 6149 & mean-per-cls \\
    \tblroworange STL-10~\cite{coates2011stl10}             & 10    & 4000  & 1000 & 8000 & accuracy \\
    \tblrowblue EuroSAT~\cite{helber2019eurosat}            & 10    & 5000  & 5000 & 5000 & accuracy \\
    \tblrowblue RESISC45~\cite{resisc45}                    & 45    & 3150  & 3150 & 25200 & accuracy \\
    Patch Camelyon~\cite{veeling2018rotation}               & 2     & 262144 & 32768 & 32768 & accuracy \\
    \midrule
    \tblroworange
    ImageNet~\cite{deng2009imagenet}     & 1000  &   1024934 & 256233 & 50000 & accuracy \\
    \tblrowgreen ImageNet A~\cite{hendrycks2019imageneta}        &  200 & N/A & N/A &  7500 & accuracy \\
    \tblrowgreen ImageNet R~\cite{hendrycks2020imagenetr}        &  200 & N/A & N/A & 30000 & accuracy \\
    \tblrowgreen ImageNet v2~\cite{recht2019imagenetv2}          & 1000 & N/A & N/A & 10000 & accuracy \\
    \tblrowgreen ImageNet Sketch~\cite{wang2019imagenetsketch}   & 1000 & N/A & N/A & 50889 & accuracy \\ 
    \bottomrule
    \end{tabularx}
\end{table}

\section{Baselines}
\label{app:baselines}
For fair evaluation, we retrained previous methods from scratch with several methods reimplemented. We also provided several system-level comparisons using pre-trained checkpoints provided by prior work. 
Below, we provide additional details on our baselines. 

\subsubsection*{Pre-trained model checkpoints}

We use publicly available checkpoints of various pre-trained models for sampling and experimental comparisons:

\begin{itemize}[topsep=0pt,itemsep=2pt,partopsep=2pt, parsep=2pt, leftmargin=10pt]

\item \textbf{SBERT~\cite{reimers2019sbert}:}
We use two checkpoints from SBERT:
\texttt{all-mpnet-base-v2} (MPNet backbone~\cite{song2020mpnet}),
and \texttt{all-MiniLM-L12-v2} (MiniLM backbone~\cite{wang2020minilm}).
Those models were used for sampling, while MPNet was also used as a frozen backbone in analysis experiments.

\item \textbf{CLIP~\cite{radford2021learning}:}
We use checkpoints available in the official Github repository\footnote{\url{https://github.com/openai/CLIP}}
for both system-level comparisons and sampling. We use the \texttt{RN50} checkpoint in the system-level comparisons to match the backbone for other models. We use the \texttt{ViT-B/32} checkpoint for sampling to provide the strongest visual sampling performance in evaluating different sampling modalities.

\item \textbf{ImageNet pre-trained model:~}
We use the checkpoints provided by \texttt{torchvision} package.\footnote{\url{https://github.com/pytorch/vision}}
For system-level comparisons, we use the ResNet-50 \texttt{IMAGENET1K\_V2} checkpoint~\cite{wightman2021resnet} as it achieves better performance than the original ResNet-50 checkpoint~\cite{he2016deep}.
We also use \texttt{ViT-B/32}~\cite{dosovitskiy2021vit} checkpoint to compare with the sampling strategy using a CLIP checkpoint.

\item \textbf{SimCLR~\cite{chen2020big}:} We use the SimCLR v2 checkpoint provided by PyTorch Lightning Bolts.\footnote{\url{https://lightning-bolts.readthedocs.io/}}
While SimCLR released some checkpoints for TensorFlow, we found that converting them to PyTorch using the recommended tools resulted in lower performance.
We use the same checkpoint for both sampling and system-level comparison.
Note that SimCLR only released models trained for 800 epochs.

\item \textbf{SimSiam~\cite{chen2020simple}:~}
We use the checkpoint trained with 512 batch size from the official Github repository\footnote{\url{https://github.com/facebookresearch/simsiam}} as it more closely matches our training setup. 

\item \textbf{MoCo~\cite{chen2021mocov3}:} We use the official checkpoint for MoCo v3.\footnote{\url{https://github.com/facebookresearch/moco-v3}} 
We use the checkpoint for the model trained for 100 epochs to match other checkpoints more closely.

\item \textbf{SwAV~\cite{caron2020unsupervised}:~} We use the official SwAV checkpoint.\footnote{\url{https://github.com/facebookresearch/swav}} 
We use the checkpoint trained for 100 epochs to match the training duration of other methods. Unlike our implementation, the full SwAV model is trained using Multi-Crop augmentation strategy. 

\end{itemize}

\subsubsection*{Retrained models }

We reimplement and retrain all baselines. When an official implementation was available, we adapted it to fit within our pipeline. 
For all models, we use a ResNet-50 backbone from torchvision with random initialization and a feature dimension of 2048 (the \texttt{fc} layer is removed). 
We use a linear layer or a multi-layer perceptron (MLP) for projection layers. Every layer except for the last is followed by batch normalization and a ReLU non-linearity. 
We describe an N-layer MLP with $N+1$ numbers depicting the input dimension for the first layer, followed by the output dimension for all layers. 

We use two forms of augmentation: SimCLR or global crop. Global crop consists of a random resized square crop with a scale of (0.5, 1.0) to an image size of $224{\times}224$. SimCLR augmentations consist of random resized square crop, color jittering, random grayscale, random horizontal flipping, and Gaussian blur. We use the same augmentation parameters as prior work~\cite{chen2020big,chen2021mocov3}. All images are normalized using ImageNet's mean and standard deviation statistics.

We provide baseline-specific details below and refer the reader to our \href{https://github.com/mbanani/lgssl}{implementation} for more details:
\begin{itemize}[topsep=4pt,itemsep=6pt,partopsep=2pt, parsep=2pt, leftmargin=10pt]

\item \textbf{SimCLR:~} We use a 3-layer MLP as a projection layer with feature dimensions (2048, 2048, 2048, 128) similar to the original paper~\cite{chen2020big}. 
We use the SimCLR loss implementation from Mu~\etal~\cite{mu2021slip}, which adapts the original loss for the distributed settings for inference and gradients. 
We use SimCLR augmentations for SimCLR and global crop augmentations for LGSimCLR. 
We experimented with mixing SimCLR augmentation and language sampled pairs and found that it results in slightly inferior performance: adding augmentations reduces the linear probe average accuracy from 78.3 to 77.9. 

\item \textbf{CLIP:~}  We use a linear projection layer to a feature dimension of 512 similar to the original paper. We use the smallest CLIP language encoder, similar to SLIP~\cite{mu2021slip}, with a feature dimension of 512 and a linear language projection layer. We use the loss implementation from SLIP~\cite{mu2021slip} but adapt it to share the loss gradients similar to the SimCLR loss. 
We use global crop augmentation for CLIP since SLIP~\cite{mu2021slip} reported that it performs better than CLIP's original center crop preprocessing. 

\item \textbf{SLIP:~}  We follow SLIP's implementation and combine the augmentations, projections, and losses from SimCLR and CLIP. We use the same language transformer as our CLIP implementation. We generate two augmented views with SimCLR augmentation for the SimCLR loss and one with global cropping for the CLIP loss. Those views are passed through their respective projections (3-layer MLP for SimCLR and linear projection for CLIP) and losses. For LGSLIP, we only use the global crop augmentation, resulting in only two augmented views and forward passes through the encoder instead of 3 for SLIP. We apply the SimCLR loss between the language-sampled image pair and the CLIP loss between only one of the images and its caption. 

\item \textbf{SimSiam:~} We follow the original SimSiam implementation and use a 3-layer MLP as our projection head (2048, 2048, 2048, 2048) and a 2-layer MLP as our prediction head (2048, 512, 2048). We use the loss formulation from the original paper. For LGSimSiam, we use the same formulation but use global crop instead of SimCLR augmentations. 

\item \textbf{SwAV. } We follow the original SwaV implementation and use a 2-layer MLP (2048, 2048, 128) as our projection head and a linear layer as our prototype head with an output dimension of 3000. The prototypes are initially frozen to improve training dynamics as suggested by the SwAV repository. We use the distributed Sinkhorn clustering implementation from the official code release. 

\item \textbf{NNCLR. } We rely on the implementation of NNCLR provided by Lightly~\cite{susmelj2020lightly} since NNCLR~\cite{dwibedi2021nnclr} did not release an implementation. Specifically, we use the memory bank implementation from Lightly and reimplement NNCLR. While our NNCLR implementation outperforms SimCLR on ImageNet, as reported in the paper, it underperforms on RedCaps. We use a 3-layer MLP (2048, 2048, 2048, 256) as our projection head and a 2-layer MLP (256, 4098, 256) as our prediction head. We also use a queue of length 16384 (equivalent to 32 batches) for the memory bank. For Language NNCLR, we also use the memory bank from Lightly, similar to DeCLIP~\cite{li2022declip}. We augment the CLIP implementation with a memory bank for the language encoder. We use a weighting of 0.8 for the CLIP loss and 0.2 for the language NNS loss, similar to DeCLIP~\cite{li2022declip}. 

\end{itemize}

\begin{table}
  \caption{\textbf{Batch Size Scaling}. The performance of both SimCLR and LGSimCLR scales with larger batch sizes with the LGSimCLR outperforming SimCLR even for larger batch sizes.}
  \label{tab:scaling_batch_size}
  \centering
  \small
  \setlength\tabcolsep{5pt}
  \resizebox{\linewidth}{!}{%
  \begin{tabularx}{\linewidth}{X cc  cc }
    \toprule
                & \multicolumn{2}{c}{SimCLR}    & \multicolumn{2}{c}{LGSimCLR} \\
                \cmidrule(l{2pt}r{2pt}){2-3} \cmidrule(l{2pt}r{2pt}){4-5}
    Batch Size   & Linear   & Fewshot      & Linear   & Fewshot \\
    \midrule 
     256                 & 67.5 & 67.2 & 77.7 & 82.3 \\
     512                 & 68.5 & 66.7 & 78.2 & 82.5 \\
    1024                 & 69.3 & 68.4 & 78.6 & 82.6 \\
    2048                 & 69.8 & 68.6 & 79.1 & 83.1 \\
    \bottomrule
  \end{tabularx}
  }
\end{table}

\section{Batch Size Scaling}
\label{app:scaling}
We explore the scaling performance of our approach for batch size. 
Prior work has shown that contrastive methods can benefit larger batch sizes~\cite{chen2020simple,chen2020big,radford2021learning}.
While we use a batch size of 512 to allow us to perform comprehensive experiments and evaluations given limited compute, we conduct a few experiments to evaluate scaling for SimCLR and LGSimCLR. 
Our results, shown in~\cref{tab:scaling_batch_size}, indicate that our approach scales with batch size and maintains our performance gains over SimCLR for larger batch sizes. Furthermore, we show in~\cref{tab:dataset_impact} that our model benefits from larger datasets. Our experiments also show that data scaling comes in two ways: training on more instances and sampling nearest neighbors from a larger pool of images. We explore this more in~\Cref{app:redcaps_shards}.

\section{Dataset Size Scaling}
\label{app:redcaps_shards}
Scaling up the dataset size not only increases the training instances,
but also \emph{broadens} the scope to sample nearest neighbors from.
In this section, we study the impact of data scaling on model performance.
First, we compare our RedCaps-trained LGSimCLR model with another model trained using a subset of RedCaps instances belonging to the year 2020.
We also train multiple LGSimCLR models using RedCaps,
each having a restricted scope of nearest neighbor sampling.
RedCaps has a natural structure to make this possible:
instances (posts) are grouped in different subreddits, across multiple years.
We expect instances within the same year to be weakly related,
and instances within a subreddit to share a consistent theme;
\eg, \textbf{\texttt{r/food}} has images of food dishes with text describing main ingredients.
We consider three restricted sampling variants of RedCaps:
Year, Subreddit, and Subreddit-Year.

We hypothesize that Subreddit sampling will limit the pool of nearest neighbor sampling to images within a similar domain resulting in higher quality neighbors.
In contrast, the Year variant will only limit the number of images to consider, reducing the probability of finding images with similar captions as neighbors.
While posts within the same year might be related to major events (\eg{} COVID-19 pandemic increased the proportion of indoor images in year 2020),
the relationship is much weaker than domain-specific subreddits.
Additionally, Subreddit-Year, which only samples the nearest neighbors from the same subreddit posted in the same year, will combine both effects.

Our results, presented in~\cref{tab:redcaps_shards}, show that domain-specific sampling improves performance. 
Meanwhile, more random sampling minimally degrades performance. 
Finally, restricting the scope of the nearest neighbor sampling is not the same as subsampling the data. 
This is shown by the higher performance of RedCaps with Year sampling compared to RedCaps-2020. 
Those results indicate two opportunities:
First, our approach can scale to very large datasets by only performing nearest-neighbor searches within subsets of the data. This is especially beneficial in some domains, such as federated learning.
Second, identifying other domain structures within the dataset can improve performance by allowing the model to sample nearest neighbor images within the same domain. 

This result indicates that our approach could scale to gigantic datasets without requiring the nearest neighbor search over the full dataset. While identifying semantically-related partitions in the datasets could improve performance, the model could perform very well by splitting randomly or using metadata information such as year or location, which might provide some relevant, although very weak, structure. 

\begin{table}
  \caption{\textbf{Impact of Sampling Scope}. LGSimCLR can still learn good features if it is restricted to only sampling from a subset of the dataset. 
  Domain-specific partitioning (\eg, subreddits) improves performance, while domain agnostic partitions (\eg, Year or Subreddit-Year) minimally degrades performance.}
  \label{tab:redcaps_shards}
  \centering
  \small
  \setlength\tabcolsep{3pt}
  \begin{tabularx}{\linewidth}{X l ccc}
    \toprule
    Dataset                     & Sampling Scope   & \#Partitions & Linear     & FewShot \\
    \midrule 
    RedCaps 2020                    & All              &    1 & 73.8 & 78.8 \\
    \midrule
    \multirow{4}{*}{RedCaps}        & All              &    1 & 78.2 & 82.5 \\
                                    & Year             &    4 & 77.2 & 80.2 \\
                                    & Subreddit        &  350 & 79.0 & 80.5 \\
                                    & Subreddit-Year   & 1391 & 77.6 & 78.2 \\
    \bottomrule
  \end{tabularx}
\end{table}

\section{Qualitative Analysis of Sampled Pairs}
\label{app:qualitative}
\begin{figure*}
  \centering
  \includegraphics[width=\linewidth]{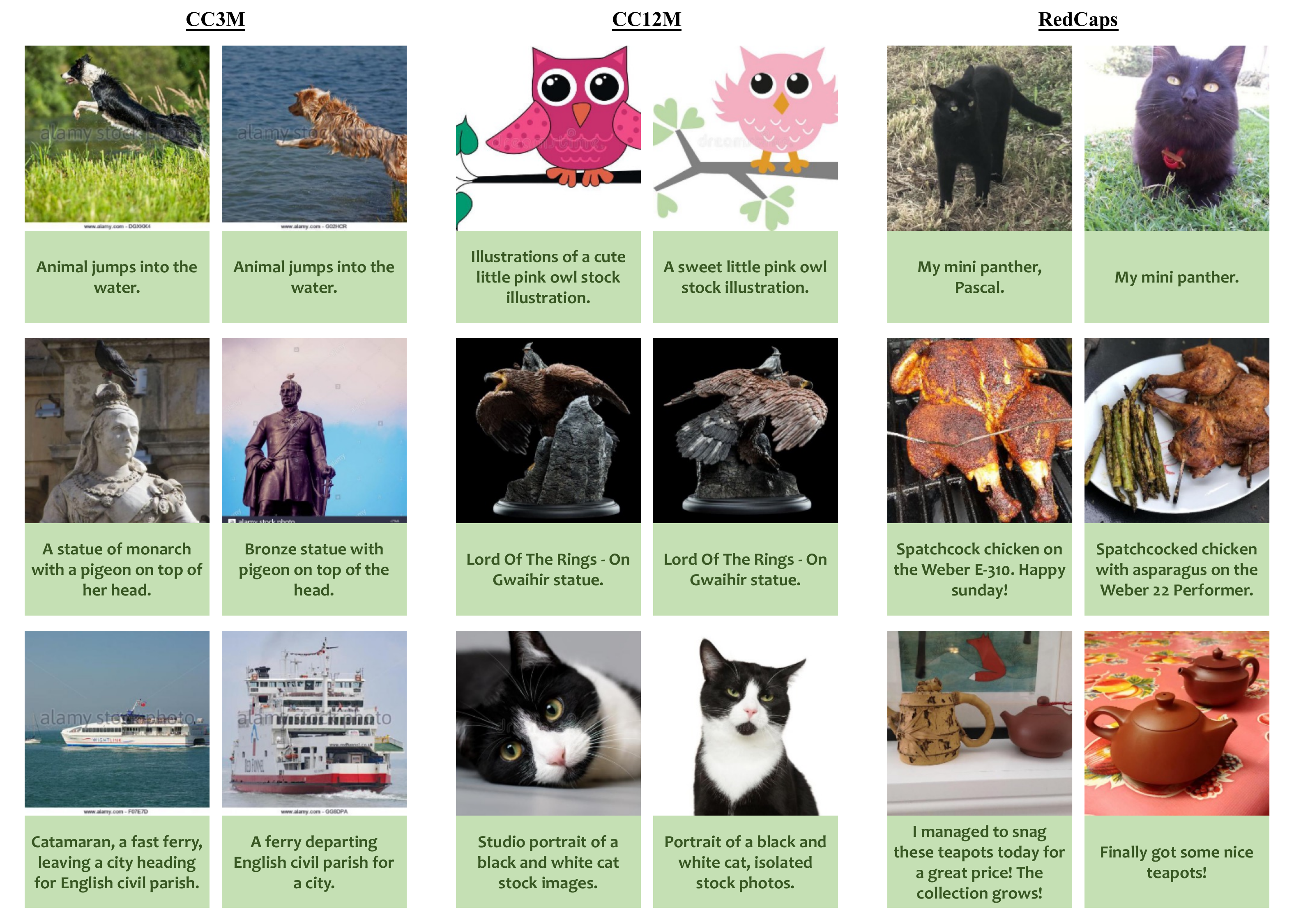}
  \caption{
  \textbf{Language sampled image pairs for the different pretraining datasets.} We show qualitative results for language guided sampling for our pretraining datasets. While the captions for conceptual captions can be generic, RedCaps captions are more natural; including longer and more natural descriptions as well as irrelevant details.  
}
  \label{fig:app_dataset_pairs}
\end{figure*}

We present language-sampled nearest neighbors for different datasets in~\cref{fig:app_dataset_pairs}.
We note that CC3M and CC12M have many stock images with slightly robotic descriptions, which is explained by how those datasets were collected.
For example, see \cref{fig:app_dataset_pairs} top row: the caption indicates `animal' instead of referring to the dog in the image.
Meanwhile, RedCaps captions can be more descriptive, referring to pet names or specific product brands or models.
Empirically, we find that RedCaps results in better performance. 

\begin{figure*}
    \centering
    \setlength{\abovecaptionskip}{4pt plus 3pt minus 2pt}
    \includegraphics[width=\linewidth]{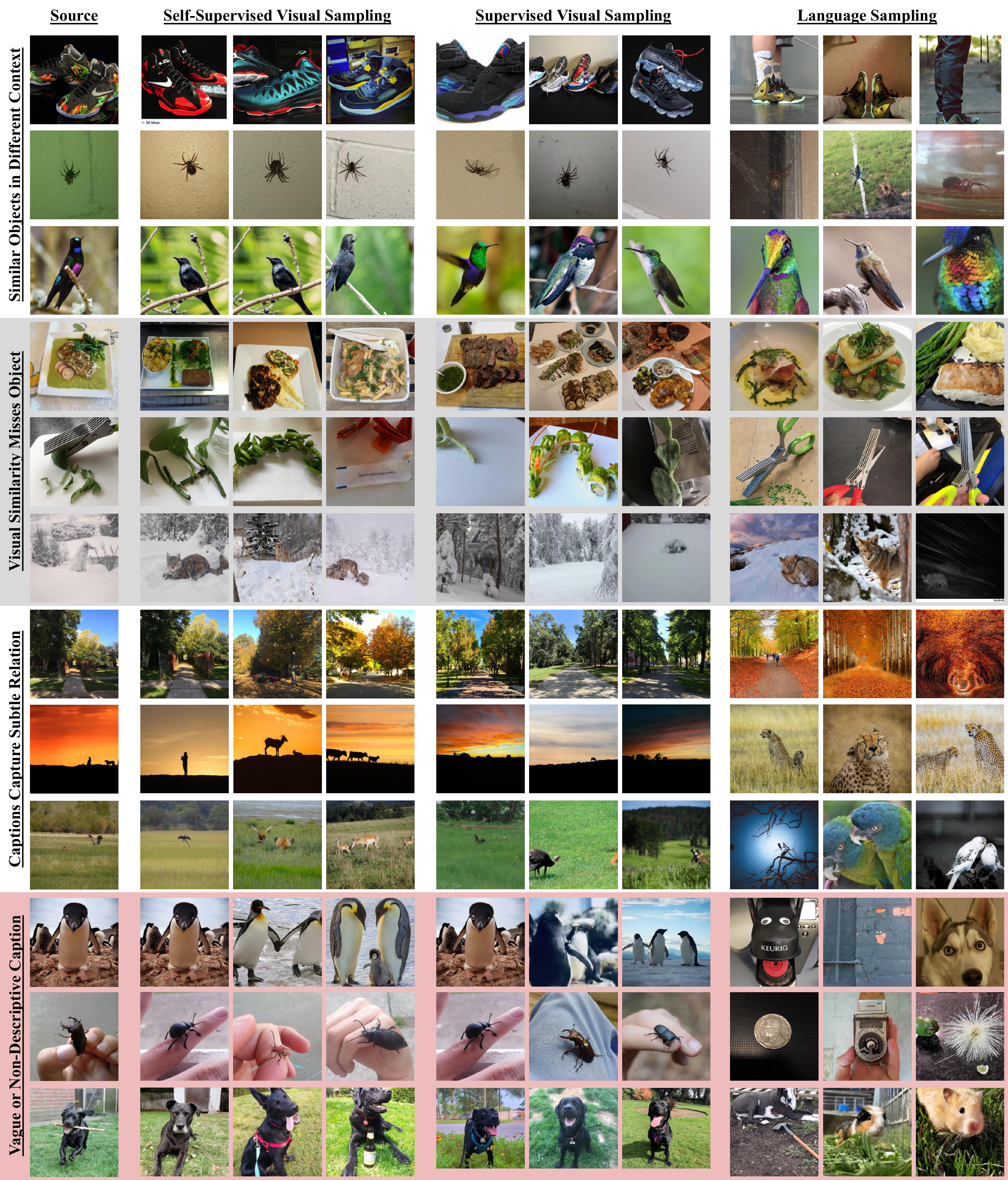}
    \caption{
    \textbf{Nearest Neighbors on RedCaps for multiple sampling options.}
  We sample the three nearest neighbors using a self-supervised visual model~\cite{chen2020big}, an ImageNet supervised model~\cite{dosovitskiy2021vit}, and a self-supervised language model~\cite{reimers2019sbert}.
  The examples are representative of some patterns we observe in the sampled pairs. 
  Language guided sampling allows us to get pairs that depict similar objects in different poses and contexts in ways that go beyond visual sampling. 
  However, sometimes the relationships depicted in the language can be too subtle. Furthermore, sometimes captions are noisy resulting in unrelated language-sampled pairs.
}
  \label{fig:app_sampled_pairs}
\end{figure*}

We also observe some repeated patterns in the types of nearest neighbor captions we get. We identify four patterns, shown in~\cref{fig:app_sampled_pairs}, and discuss them below:

\llsparagraph{Similar objects in different contexts:}
The first set of results shows examples where language-guided sampling results in diverse images depicting the same concept.
For example, while pairs sampled using visual models (regardless of whether they are self-supervised or supervised) depict shoes on their own, language samples three images of the same shoe model in very different contexts.
The third row also depicts hummingbirds in very different poses.
At the same time, self-supervised models provide three birds on a branch, and supervised models provide three hummingbirds taken in similar poses as the source image.

\llsparagraph{Visual similarity misses the object:}
The second set shows examples where visual similarity misses the salient object in the image.
The fourth row is a halibut dish with vegetables.
Visual sampling results in other dishes pictured from the top, while language sampling gives us three other halibut dishes with vegetables that look different from the source image.
Rows 5 and 6 show examples where visual sampling focused on the overall appearance and missing the herb scissors (row 5) and coyote (row 6). Self-supervised models provide the nearest neighbors with animals in the snow, but different animals like a lynx or a dog. 

\llsparagraph{Captions capture subtle relationships:}
The third set shows examples where the language captures subtle relationships. Can you guess what the captions were? In row 7, the source image was captioned ``\textit{itap of a tunnel created by the autumn leaves}.'' Visual similarity focuses on the trees, while language similarity results in images depicting autumn more clearly. In row 8, the source caption mentions a cheetah which can be seen at the right corner of the source image, but the overall sunset appearance results in different sets of visual nearest neighbors. Finally, the caption for row 9 mentions a mating ritual between birds. This element is captured by language guidance, while visual similarity retrieves images of animals in the grass. These results suggest that conditioning the model similarity on the caption could result in a better-posed learning problem. 

\llsparagraph{Vague Captions: }
Those examples show cases where the caption is very vague or unrelated to the image content, resulting in odd nearest neighbors in the language space. Can you guess the captions from the nearest neighbor images? Answers are in the footnote.\footnote{Source Image Captions: \\ 
row 10: ``\textit{Built-in googly eyes.}''	\\ 
row 11: ``\textit{I found this today. Anyone knows what it is?}''	\\ 
row 12: ``\textit{Cinda having fun in the garden!}''}
The caption of row 10 refers to the appearance of the eyes of the penguin, but since the ``googly eyes'' can also refer to a small toy, it retrieves images of that toy being used on a coffee machine and a wall. 
In row 11, the caption asks what the object is, but this is independent of the object. This results in language retrievals with miscellaneous objects, while visual retrievals return other insects. 
Finally, row 12 shows a case where the retrieval uses the dog's name in some context, resulting in the retrieval of other pets playing in gardens. These cases represent limitations of language sampling that might result in poor learning. However, since the core issue arises from misalignment or vagueness in the caption, it is a limitation shared by any model that uses captions and images.

\section{Additional Results}
\label{app:results}
Due to space limitations, we only report average performance for several methods in the main paper. Here, we report the complete performance breakdown for all methods on linear probe in~\cref{tab:app_linear_probe} and fewshot classification~\cref{tab:app_fewshot}. For fewshot classification, we also report the 95\% confidence interval as a subscript. 

We also evaluate all approaches on several ImageNet evaluation benchmarks. We use the same evaluation setups described in~\Cref{app:evaluation_tasks} and report results in~\cref{tab:app_imageneteval}
Specifically, we train on the ImageNet train set and evaluate on the ImageNet validation set~\cite{deng2009imagenet}, and several alternative ImageNet test splits that assess robustness.
ImageNet A(dverserial)~\cite{hendrycks2019imageneta}, ImageNet R(enditions)~\cite{hendrycks2020imagenetr}, ImageNet v2~\cite{recht2019imagenetv2}, and ImageNet Sketch~\cite{wang2019imagenetsketch}. 
We observe similar performance trends for the RedCaps-trained models, with LGSimCLR outperforming all baselines, LGSLIP outperforming LGSimCLR, and training on larger datasets or with larger batch sizes improving performance. 
We also note that the difference in performance between our models and the ImageNet pre-trained checkpoints is larger due to the smaller domain shift they experience from ImageNet training.

\begin{table*}
  \newcommand{\rowheader}{\rowcolor{Gray!20}}
  \centering
    \caption{
  \textbf{Linear Probe Evaluations.} We report the linear probe classification performance of all baselines and models. Models are grouped by experiment. }
  \label{tab:app_linear_probe}
  \setlength\tabcolsep{2.5pt}
\resizebox{\linewidth}{!}{%
  \begin{tabular}{llc cccccccccccccccc}
    \toprule
    \textbf{Model} &
    \textbf{Dataset} &
    \textbf{Sampling Space} &
    \rotatebox[origin=lb]{90}{\bf {Food-101}} &
    \rotatebox[origin=lb]{90}{\bf {CIFAR-10}} &
    \rotatebox[origin=lb]{90}{\bf {CIFAR-100}} &
    \rotatebox[origin=lb]{90}{\bf {CUB}} &
    \rotatebox[origin=lb]{90}{\bf {SUN397}} &
    \rotatebox[origin=lb]{90}{\bf {Cars}} &
    \rotatebox[origin=lb]{90}{\bf {Aircraft}} &
    \rotatebox[origin=lb]{90}{\bf {DTD}} &
    \rotatebox[origin=lb]{90}{\bf {Pets}} &
    \rotatebox[origin=lb]{90}{\bf {Caltech-101~}} &
    \rotatebox[origin=lb]{90}{\bf {Flowers}} &
    \rotatebox[origin=lb]{90}{\bf {STL-10}} &
    \rotatebox[origin=lb]{90}{\bf {EuroSAT}} &
    \rotatebox[origin=lb]{90}{\bf {RESISC45}} &
    \rotatebox[origin=lb]{90}{\bf {PCAM}} &
    \rotatebox[origin=lb]{0}{\bf {Avg.}} \\
    \midrule
\rowheader \multicolumn{19}{l}{\textit{\textbf{Pre-trained Checkpoints}}} \\
Supervised~\cite{wightman2021resnet}                       & ImageNet             & -                              &  71.0 &  93.2 &  77.0 &  68.4 &  63.0 &  48.7 &  41.0 &  73.0 &  92.6 &  91.9 &  88.3 &  98.2 &  95.8 &  85.3 &  82.4 & 78.0 \\
SimSiam~\cite{chen2021exploring}                        & ImageNet             & -                              &  70.6 &  92.2 &  74.9 &  47.1 &   0.3 &  52.4 &  52.6 &  74.4 &  83.3 &  89.3 &  91.8 &  95.9 &  96.7 &  86.4 &  85.2 & 72.9 \\
MoCo v3~\cite{chen2021mocov3}                        & ImageNet             & -                              &  71.4 &  93.3 &  77.9 &  51.5 &  60.4 &  52.4 &  53.0 &  73.6 &  85.8 &  90.4 &  92.1 &  96.8 &  96.3 &  85.0 &  85.0 & 77.7 \\
SwAV~\cite{caron2020unsupervised}                           & ImageNet             & -                              &  72.8 &  93.0 &  77.5 &  48.8 &  63.2 &  55.5 &  52.7 &  77.2 &  84.5 &  89.9 &  93.4 &  97.2 &  96.7 &  86.9 &  83.7 & 78.2 \\
SimCLR~\cite{chen2020big}                         & ImageNet             & -                              &  71.4 &  91.3 &  73.9 &  44.3 &  60.3 &  44.6 &  46.7 &  74.9 &  83.9 &  87.4 &  90.2 &  96.2 &  95.9 &  84.4 &  85.1 & 75.4 \\
CLIP~\cite{radford2021learning}                           & CLIP (400M)          & -                              &  86.4 &  88.7 &  70.2 &  69.8 &  72.5 &  78.4 &  49.4 &  76.3 &  88.0 &  88.9 &  96.1 &  97.2 &  94.7 &  87.9 &  82.7 & 81.8 \\
\midrule \rowheader \multicolumn{19}{l}{\textit{\textbf{RedCaps-trained Baselines}}} \\
SwAV                           & RedCaps              & -                              &  63.6 &  81.3 &  57.5 &  21.6 &  47.5 &  22.9 &  35.4 &  68.1 &  61.1 &  70.5 &  78.0 &  87.7 &  94.3 &  79.9 &  84.3 & 63.6 \\
SimSiam                        & RedCaps              & -                              &  64.1 &  79.9 &  56.1 &  28.2 &  48.3 &  29.5 &  41.2 &  66.2 &  69.1 &  73.6 &  83.6 &  85.7 &  94.4 &  82.1 &  83.3 & 65.7 \\
SimCLR                         & RedCaps              & -                              &  69.0 &  82.9 &  61.6 &  30.6 &  52.6 &  33.7 &  43.7 &  69.8 &  70.5 &  74.1 &  86.9 &  88.0 &  95.4 &  84.6 &  84.4 & 68.5 \\
Visual NNCLR                   & RedCaps              & -                              &  65.4 &  82.8 &  60.2 &  26.6 &  50.0 &  26.6 &  40.9 &  68.0 &  65.2 &  75.4 &  83.5 &  88.5 &  95.3 &  82.2 &  83.8 & 66.3 \\
CLIP                           & RedCaps              & -                              &  80.9 &  84.7 &  62.7 &  50.4 &  57.4 &  45.8 &  36.7 &  67.6 &  79.8 &  84.0 &  91.0 &  93.5 &  93.9 &  82.2 &  82.6 & 72.9 \\
CLIP (SBERT Encoder)           & RedCaps              & -                              &  80.5 &  81.3 &  59.4 &  50.6 &  56.9 &  45.9 &  35.7 &  69.1 &  76.7 &  81.7 &  90.2 &  93.6 &  92.9 &  81.1 &  81.3 & 71.8 \\
Language NNCLR                 & RedCaps              & -                              &  81.2 &  83.1 &  61.9 &  48.6 &  56.5 &  45.1 &  37.2 &  68.8 &  78.1 &  82.0 &  90.2 &  93.4 &  92.5 &  81.1 &  80.7 & 72.0 \\
SLIP                           & RedCaps              & -                              &  77.7 &  87.2 &  67.0 &  42.4 &  58.1 &  48.7 &  45.2 &  72.3 &  79.5 &  82.7 &  92.1 &  92.7 &  95.6 &  85.5 &  83.4 & 74.0 \\
\midrule \rowheader \multicolumn{19}{l}{\textit{\textbf{Sampling Space - Language}}} \\
LGSimCLR                       & RedCaps              & SBERT (MiniLM)                 &  83.2 &  88.0 &  69.3 &  60.4 &  59.7 &  64.0 &  54.0 &  72.7 &  82.6 &  88.5 &  95.7 &  94.1 &  96.4 &  88.1 &  82.2 & 78.6 \\
LGSimCLR                       & RedCaps              & CLIP (400M)                    &  83.3 &  87.6 &  68.9 &  60.1 &  59.9 &  62.9 &  53.7 &  70.5 &  82.6 &  88.7 &  95.6 &  94.3 &  96.2 &  88.2 &  82.0 & 78.3 \\
LGSimCLR                       & RedCaps              & CLIP (RedCaps)                 &  83.7 &  88.0 &  67.8 &  59.6 &  60.7 &  60.8 &  53.7 &  71.4 &  82.4 &  89.1 &  95.9 &  93.8 &  96.1 &  88.4 &  82.7 & 78.3 \\
LGSimCLR                       & RedCaps              & FastText BoW                   &  80.8 &  85.5 &  66.7 &  54.2 &  58.7 &  56.6 &  51.1 &  69.9 &  78.3 &  88.0 &  94.4 &  92.5 &  96.2 &  87.7 &  81.5 & 76.1 \\
\midrule \rowheader \multicolumn{19}{l}{\textit{\textbf{Sampling Space - Visual}}} \\
LGSimCLR                       & RedCaps              & ImageNet Supervised            &  75.7 &  92.2 &  75.4 &  57.5 &  60.2 &  53.7 &  52.2 &  71.7 &  90.3 &  90.2 &  93.1 &  95.5 &  96.8 &  87.7 &  83.0 & 78.3 \\
LGSimCLR                       & RedCaps              & SimCLR                         &  71.4 &  87.0 &  67.5 &  36.8 &  57.9 &  41.8 &  46.3 &  74.2 &  82.8 &  82.6 &  90.7 &  93.4 &  95.7 &  85.2 &  83.2 & 73.1 \\
LGSimCLR                       & RedCaps              & CLIP (400M)                    &  83.6 &  90.7 &  72.1 &  58.3 &  62.5 &  59.2 &  51.3 &  75.5 &  88.7 &  90.3 &  95.2 &  95.4 &  96.2 &  88.6 &  82.9 & 79.4 \\
\midrule \rowheader \multicolumn{19}{l}{\textit{\textbf{Sampling Scope}}} \\
LGSimCLR                       & RedCaps              & SBERT - Year                   &  82.6 &  85.8 &  66.1 &  58.1 &  59.0 &  57.1 &  52.8 &  71.9 &  80.7 &  88.1 &  95.6 &  93.1 &  96.1 &  88.0 &  82.8 & 77.2 \\
LGSimCLR                       & RedCaps              & SBERT - Sub-Year               &  82.4 &  86.8 &  66.9 &  56.4 &  59.5 &  54.5 &  51.4 &  72.1 &  89.0 &  89.8 &  94.9 &  95.2 &  95.8 &  88.1 &  81.9 & 77.6 \\
LGSimCLR                       & RedCaps              & SBERT - Sub                    &  83.2 &  88.7 &  69.5 &  60.4 &  59.9 &  60.0 &  53.0 &  72.4 &  89.7 &  90.3 &  96.2 &  94.8 &  96.2 &  88.4 &  82.2 & 79.0 \\
\midrule \rowheader \multicolumn{19}{l}{\textit{\textbf{Pre-training Datasets}}} \\
LGSimCLR                       & CC3M                 & SBERT (MPNet)                  &  64.4 &  84.6 &  65.4 &  44.4 &  59.1 &  41.9 &  46.8 &  66.0 &  70.7 &  83.9 &  91.2 &  91.6 &  95.6 &  86.1 &  80.5 & 71.5 \\
LGSimCLR                       & CC12M                & SBERT (MPNet)                  &  73.4 &  88.6 &  70.1 &  50.4 &  66.0 &  58.7 &  52.4 &  72.6 &  79.0 &  88.3 &  92.6 &  94.5 &  95.6 &  87.5 &  81.6 & 76.8 \\
LGSimCLR                       & RedCaps 2020         & SBERT (MPNet)                  &  77.8 &  84.3 &  64.5 &  53.9 &  53.9 &  51.7 &  48.1 &  66.4 &  76.2 &  83.9 &  93.9 &  89.9 &  95.4 &  86.4 &  81.4 & 73.8 \\
\midrule \rowheader \multicolumn{19}{l}{\textit{\textbf{Batch Size Scaling}}} \\
SimCLR (256)                   & RedCaps              & -                              &  67.1 &  83.1 &  60.5 &  28.5 &  51.0 &  32.5 &  42.4 &  70.0 &  68.3 &  73.9 &  85.8 &  86.5 &  96.2 &  84.2 &  83.1 & 67.5 \\
SimCLR (1024)                  & RedCaps              & -                              &  70.0 &  84.4 &  62.8 &  31.9 &  52.4 &  35.8 &  44.2 &  70.9 &  72.7 &  74.7 &  87.9 &  88.3 &  95.6 &  84.8 &  83.3 & 69.3 \\
SimCLR (2048)                  & RedCaps              & -                              &  70.4 &  83.9 &  62.6 &  32.5 &  53.3 &  36.7 &  44.9 &  70.9 &  73.1 &  75.5 &  88.1 &  88.8 &  96.5 &  85.1 &  84.2 & 69.8 \\
LGSimCLR (256)                 & RedCaps              & SBERT (MPNet)                  &  82.9 &  87.3 &  68.0 &  58.7 &  60.2 &  58.2 &  52.6 &  73.2 &  81.1 &  88.2 &  95.2 &  94.0 &  96.1 &  87.7 &  81.8 & 77.7 \\
LGSimCLR (1024)                & RedCaps              & SBERT (MPNet)                  &  83.7 &  87.1 &  68.1 &  62.0 &  60.7 &  63.4 &  53.7 &  73.4 &  80.8 &  89.8 &  95.7 &  93.7 &  95.7 &  88.3 &  82.6 & 78.6 \\
LGSimCLR (2048)                & RedCaps              & SBERT (MPNet)                  &  84.2 &  88.2 &  69.1 &  63.2 &  60.9 &  65.2 &  55.5 &  71.4 &  81.7 &  89.7 &  96.0 &  94.4 &  96.3 &  88.5 &  82.1 & 79.1 \\
\midrule \rowheader \multicolumn{19}{l}{\textit{\textbf{Alternative Formulations}}} \\
LGSimCLR                       & RedCaps              & SBERT (MPNet)                  &  83.2 &  87.8 &  69.0 &  59.3 &  60.3 &  62.3 &  53.4 &  71.2 &  81.8 &  89.4 &  95.9 &  94.0 &  95.6 &  88.0 &  81.1 & 78.2 \\
LGSimSiam                      & RedCaps              & SBERT (MPNet)                  &  73.8 &  83.4 &  62.6 &  40.6 &  54.6 &  41.1 &  47.3 &  68.6 &  66.5 &  85.2 &  90.3 &  90.8 &  95.7 &  85.6 &  81.3 & 71.2 \\
LGSLIP                         & RedCaps              & SBERT (MPNet)                  &  84.5 &  87.4 &  69.2 &  60.7 &  62.3 &  62.2 &  52.5 &  73.1 &  83.1 &  90.2 &  96.3 &  94.8 &  95.3 &  88.4 &  82.7 & 78.8 \\
\bottomrule
\end{tabular}
}
\end{table*}
\begin{table*}
  \centering
  \newcommand{\rowheader}{\rowcolor{Gray!20}}
    \caption{
  \textbf{Few-Shot Evaluations.} We report the 5-way, 5-shot classification performance of all baselines and models. Models are grouped by experiment. The subscript reports the 95\% confidence interval in prediction across 5000 episodes.  }
  \label{tab:app_fewshot}
  \setlength\tabcolsep{2.5pt}
\resizebox{\linewidth}{!}{%
  \begin{tabular}{llc ccccccccccccccc}
    \toprule
    \textbf{Model} &
    \textbf{Dataset} &
    \textbf{Sampling Space} &
    \rotatebox[origin=lb]{90}{\bf {Food-101}} &
    \rotatebox[origin=lb]{90}{\bf {CIFAR-10}} &
    \rotatebox[origin=lb]{90}{\bf {CIFAR-100}} &
    \rotatebox[origin=lb]{90}{\bf {CUB}} &
    \rotatebox[origin=lb]{90}{\bf {SUN397}} &
    \rotatebox[origin=lb]{90}{\bf {Cars}} &
    \rotatebox[origin=lb]{90}{\bf {Aircraft}} &
    \rotatebox[origin=lb]{90}{\bf {DTD}} &
    \rotatebox[origin=lb]{90}{\bf {Pets}} &
    \rotatebox[origin=lb]{90}{\bf {Caltech-101~}} &
    \rotatebox[origin=lb]{90}{\bf {Flowers}} &
    \rotatebox[origin=lb]{90}{\bf {STL-10}} &
    \rotatebox[origin=lb]{90}{\bf {EuroSAT}} &
    \rotatebox[origin=lb]{90}{\bf {RESISC45}} &
    \rotatebox[origin=lb]{0}{\bf {Avg.}} \\
    \midrule
\rowheader \multicolumn{18}{l}{\textit{\textbf{Pre-trained Checkpoints}}} \\
Supervised~\cite{wightman2021resnet}                       & ImageNet             & -                              & \fewshot{81.6}{0.3} & \fewshot{84.1}{0.2} & \fewshot{87.8}{0.2} & \fewshot{91.9}{0.2} & \fewshot{95.0}{0.1} & \fewshot{75.7}{0.3} & \fewshot{53.1}{0.4} & \fewshot{80.4}{0.3} & \fewshot{97.6}{0.1} & \fewshot{97.4}{0.1} & \fewshot{91.4}{0.2} & \fewshot{95.2}{0.1} & \fewshot{83.6}{0.2} & \fewshot{85.2}{0.3} & 85.7\\
SimSiam~\cite{chen2021exploring}                        & ImageNet             & -                              & \fewshot{70.5}{0.3} & \fewshot{77.5}{0.3} & \fewshot{82.0}{0.3} & \fewshot{67.4}{0.4} & \fewshot{92.1}{0.2} & \fewshot{51.9}{0.3} & \fewshot{43.7}{0.4} & \fewshot{81.8}{0.3} & \fewshot{86.7}{0.3} & \fewshot{94.9}{0.2} & \fewshot{93.7}{0.2} & \fewshot{89.7}{0.2} & \fewshot{87.7}{0.2} & \fewshot{82.2}{0.3} & 78.7\\
MoCo v3~\cite{chen2021mocov3}                        & ImageNet             & -                              & \fewshot{72.7}{0.3} & \fewshot{82.3}{0.2} & \fewshot{84.9}{0.3} & \fewshot{74.7}{0.3} & \fewshot{92.5}{0.2} & \fewshot{52.0}{0.4} & \fewshot{42.4}{0.4} & \fewshot{80.7}{0.3} & \fewshot{89.1}{0.2} & \fewshot{95.8}{0.1} & \fewshot{93.6}{0.2} & \fewshot{91.5}{0.2} & \fewshot{86.3}{0.2} & \fewshot{83.0}{0.3} & 80.1\\
SwAV~\cite{caron2020unsupervised}                           & ImageNet             & -                              & \fewshot{68.3}{0.3} & \fewshot{78.1}{0.3} & \fewshot{82.1}{0.3} & \fewshot{65.4}{0.4} & \fewshot{93.7}{0.2} & \fewshot{52.7}{0.4} & \fewshot{40.3}{0.4} & \fewshot{83.8}{0.2} & \fewshot{83.8}{0.3} & \fewshot{94.5}{0.2} & \fewshot{93.4}{0.2} & \fewshot{91.2}{0.2} & \fewshot{88.0}{0.2} & \fewshot{83.8}{0.3} & 78.5\\
SimCLR~\cite{chen2020big}                         & ImageNet             & -                              & \fewshot{70.0}{0.3} & \fewshot{76.9}{0.3} & \fewshot{80.9}{0.3} & \fewshot{67.5}{0.4} & \fewshot{92.5}{0.2} & \fewshot{51.9}{0.3} & \fewshot{42.1}{0.4} & \fewshot{82.2}{0.3} & \fewshot{85.0}{0.3} & \fewshot{93.0}{0.2} & \fewshot{90.3}{0.2} & \fewshot{88.8}{0.2} & \fewshot{83.6}{0.3} & \fewshot{78.5}{0.3} & 77.4\\
CLIP~\cite{radford2021learning}                           & CLIP (400M)          & -                              & \fewshot{92.1}{0.2} & \fewshot{76.3}{0.3} & \fewshot{79.2}{0.3} & \fewshot{92.9}{0.2} & \fewshot{96.9}{0.1} & \fewshot{93.3}{0.2} & \fewshot{73.2}{0.4} & \fewshot{81.8}{0.3} & \fewshot{86.1}{0.3} & \fewshot{95.9}{0.1} & \fewshot{97.9}{0.1} & \fewshot{95.6}{0.1} & \fewshot{77.5}{0.3} & \fewshot{90.5}{0.2} & 87.8\\
\midrule \rowheader \multicolumn{18}{l}{\textit{\textbf{RedCaps-trained Baselines}}} \\
SwAV                           & RedCaps              & -                              & \fewshot{64.5}{0.4} & \fewshot{54.0}{0.3} & \fewshot{61.8}{0.3} & \fewshot{45.8}{0.4} & \fewshot{84.9}{0.3} & \fewshot{36.5}{0.3} & \fewshot{34.1}{0.3} & \fewshot{74.8}{0.3} & \fewshot{66.5}{0.4} & \fewshot{78.1}{0.3} & \fewshot{75.5}{0.3} & \fewshot{72.6}{0.3} & \fewshot{80.4}{0.3} & \fewshot{72.9}{0.4} & 64.5\\
SimSiam                        & RedCaps              & -                              & \fewshot{63.9}{0.3} & \fewshot{49.9}{0.3} & \fewshot{57.2}{0.3} & \fewshot{49.5}{0.4} & \fewshot{84.5}{0.3} & \fewshot{39.3}{0.3} & \fewshot{37.9}{0.3} & \fewshot{75.7}{0.3} & \fewshot{67.8}{0.4} & \fewshot{79.7}{0.3} & \fewshot{81.5}{0.3} & \fewshot{69.6}{0.3} & \fewshot{80.6}{0.3} & \fewshot{79.4}{0.3} & 65.5\\
SimCLR                         & RedCaps              & -                              & \fewshot{66.9}{0.3} & \fewshot{45.7}{0.3} & \fewshot{51.0}{0.3} & \fewshot{51.5}{0.4} & \fewshot{87.1}{0.2} & \fewshot{44.0}{0.3} & \fewshot{38.4}{0.3} & \fewshot{77.6}{0.3} & \fewshot{70.1}{0.3} & \fewshot{80.0}{0.3} & \fewshot{86.9}{0.2} & \fewshot{69.6}{0.3} & \fewshot{83.5}{0.3} & \fewshot{81.3}{0.3} & 66.7\\
Visual NNCLR                   & RedCaps              & -                              & \fewshot{65.6}{0.3} & \fewshot{54.1}{0.3} & \fewshot{61.7}{0.3} & \fewshot{45.8}{0.3} & \fewshot{85.3}{0.3} & \fewshot{37.9}{0.3} & \fewshot{34.9}{0.3} & \fewshot{75.2}{0.3} & \fewshot{67.3}{0.4} & \fewshot{81.1}{0.3} & \fewshot{75.4}{0.3} & \fewshot{74.3}{0.3} & \fewshot{83.6}{0.3} & \fewshot{76.7}{0.3} & 65.6\\
CLIP                           & RedCaps              & -                              & \fewshot{88.9}{0.2} & \fewshot{64.6}{0.3} & \fewshot{73.1}{0.3} & \fewshot{78.3}{0.3} & \fewshot{90.9}{0.2} & \fewshot{69.7}{0.3} & \fewshot{40.7}{0.3} & \fewshot{75.7}{0.3} & \fewshot{77.5}{0.3} & \fewshot{91.6}{0.2} & \fewshot{94.7}{0.2} & \fewshot{89.8}{0.2} & \fewshot{75.3}{0.3} & \fewshot{74.8}{0.3} & 77.5\\
CLIP (SBERT Encoder)           & RedCaps              & -                              & \fewshot{89.9}{0.2} & \fewshot{59.9}{0.3} & \fewshot{67.9}{0.3} & \fewshot{83.2}{0.3} & \fewshot{91.1}{0.2} & \fewshot{70.2}{0.3} & \fewshot{41.0}{0.3} & \fewshot{75.0}{0.3} & \fewshot{79.4}{0.3} & \fewshot{91.2}{0.2} & \fewshot{94.5}{0.2} & \fewshot{89.4}{0.2} & \fewshot{72.3}{0.3} & \fewshot{74.9}{0.3} & 77.1\\
Language NNCLR                 & RedCaps              & -                              & \fewshot{89.3}{0.2} & \fewshot{65.3}{0.3} & \fewshot{73.4}{0.3} & \fewshot{78.6}{0.3} & \fewshot{90.8}{0.2} & \fewshot{68.4}{0.3} & \fewshot{40.4}{0.3} & \fewshot{75.2}{0.3} & \fewshot{78.8}{0.3} & \fewshot{90.9}{0.2} & \fewshot{94.3}{0.2} & \fewshot{89.6}{0.2} & \fewshot{75.2}{0.3} & \fewshot{71.9}{0.3} & 77.3\\
SLIP                           & RedCaps              & -                              & \fewshot{81.5}{0.3} & \fewshot{63.5}{0.3} & \fewshot{70.8}{0.3} & \fewshot{63.1}{0.4} & \fewshot{91.3}{0.2} & \fewshot{62.9}{0.3} & \fewshot{42.1}{0.4} & \fewshot{79.6}{0.3} & \fewshot{76.4}{0.3} & \fewshot{88.4}{0.2} & \fewshot{92.2}{0.2} & \fewshot{83.4}{0.2} & \fewshot{82.7}{0.3} & \fewshot{80.8}{0.3} & 75.6\\
\midrule \rowheader \multicolumn{18}{l}{\textit{\textbf{Sampling Space - Language}}} \\
LGSimCLR                       & RedCaps              & SBERT (MiniLM)                 & \fewshot{90.4}{0.2} & \fewshot{67.1}{0.3} & \fewshot{76.7}{0.3} & \fewshot{83.9}{0.3} & \fewshot{92.7}{0.2} & \fewshot{79.2}{0.3} & \fewshot{52.1}{0.4} & \fewshot{81.2}{0.3} & \fewshot{86.2}{0.3} & \fewshot{95.5}{0.1} & \fewshot{97.6}{0.1} & \fewshot{87.4}{0.2} & \fewshot{86.9}{0.2} & \fewshot{89.0}{0.2} & 83.3\\
LGSimCLR                       & RedCaps              & CLIP (400M)                    & \fewshot{90.7}{0.2} & \fewshot{65.8}{0.3} & \fewshot{75.6}{0.3} & \fewshot{83.8}{0.3} & \fewshot{92.8}{0.2} & \fewshot{80.9}{0.3} & \fewshot{52.0}{0.4} & \fewshot{81.4}{0.3} & \fewshot{85.6}{0.3} & \fewshot{95.5}{0.1} & \fewshot{97.5}{0.1} & \fewshot{87.3}{0.2} & \fewshot{84.8}{0.2} & \fewshot{89.3}{0.2} & 83.1\\
LGSimCLR                       & RedCaps              & CLIP (RedCaps)                 & \fewshot{90.4}{0.2} & \fewshot{64.8}{0.3} & \fewshot{75.3}{0.3} & \fewshot{82.2}{0.3} & \fewshot{92.8}{0.2} & \fewshot{76.6}{0.3} & \fewshot{50.4}{0.4} & \fewshot{81.3}{0.3} & \fewshot{84.6}{0.3} & \fewshot{95.2}{0.2} & \fewshot{97.7}{0.1} & \fewshot{86.9}{0.2} & \fewshot{86.5}{0.2} & \fewshot{89.1}{0.2} & 82.4\\
LGSimCLR                       & RedCaps              & FastText BoW                   & \fewshot{88.4}{0.2} & \fewshot{62.1}{0.3} & \fewshot{73.7}{0.3} & \fewshot{79.3}{0.3} & \fewshot{92.2}{0.2} & \fewshot{74.0}{0.3} & \fewshot{52.5}{0.4} & \fewshot{79.4}{0.3} & \fewshot{82.7}{0.3} & \fewshot{94.3}{0.2} & \fewshot{97.5}{0.1} & \fewshot{83.0}{0.2} & \fewshot{85.4}{0.2} & \fewshot{88.5}{0.2} & 80.9\\
\midrule \rowheader \multicolumn{18}{l}{\textit{\textbf{Sampling Space - Visual}}}  \\
LGSimCLR                       & RedCaps              & ImageNet Supervised            & \fewshot{79.6}{0.3} & \fewshot{75.6}{0.3} & \fewshot{83.0}{0.3} & \fewshot{76.6}{0.3} & \fewshot{92.5}{0.2} & \fewshot{64.6}{0.4} & \fewshot{46.1}{0.4} & \fewshot{80.7}{0.3} & \fewshot{94.3}{0.2} & \fewshot{96.3}{0.1} & \fewshot{94.8}{0.2} & \fewshot{87.4}{0.2} & \fewshot{86.4}{0.2} & \fewshot{87.3}{0.2} & 81.8\\
LGSimCLR                       & RedCaps              & SimCLR                         & \fewshot{72.0}{0.3} & \fewshot{62.9}{0.3} & \fewshot{71.9}{0.3} & \fewshot{58.9}{0.4} & \fewshot{90.8}{0.2} & \fewshot{51.3}{0.3} & \fewshot{38.7}{0.3} & \fewshot{81.8}{0.3} & \fewshot{86.4}{0.3} & \fewshot{91.3}{0.2} & \fewshot{90.7}{0.2} & \fewshot{83.8}{0.2} & \fewshot{85.5}{0.2} & \fewshot{78.6}{0.3} & 74.6\\
LGSimCLR                       & RedCaps              & CLIP (400M)                    & \fewshot{88.8}{0.2} & \fewshot{72.5}{0.3} & \fewshot{79.7}{0.3} & \fewshot{77.6}{0.3} & \fewshot{93.1}{0.2} & \fewshot{73.3}{0.3} & \fewshot{45.6}{0.4} & \fewshot{82.2}{0.3} & \fewshot{90.9}{0.2} & \fewshot{94.6}{0.2} & \fewshot{96.3}{0.1} & \fewshot{89.2}{0.2} & \fewshot{84.6}{0.2} & \fewshot{87.4}{0.2} & 82.6\\
\midrule \rowheader \multicolumn{18}{l}{\textit{\textbf{Sampling Scope}}} \\
LGSimCLR                       & RedCaps              & SBERT - Year                   & \fewshot{89.7}{0.2} & \fewshot{61.2}{0.3} & \fewshot{72.2}{0.3} & \fewshot{81.2}{0.3} & \fewshot{92.0}{0.2} & \fewshot{71.9}{0.3} & \fewshot{48.4}{0.4} & \fewshot{80.1}{0.3} & \fewshot{79.3}{0.3} & \fewshot{93.8}{0.2} & \fewshot{97.4}{0.1} & \fewshot{84.5}{0.2} & \fewshot{84.0}{0.2} & \fewshot{87.8}{0.2} & 80.2\\
LGSimCLR                       & RedCaps              & SBERT - Sub-Year               & \fewshot{88.0}{0.2} & \fewshot{59.0}{0.3} & \fewshot{66.6}{0.3} & \fewshot{76.5}{0.3} & \fewshot{90.6}{0.2} & \fewshot{63.6}{0.4} & \fewshot{45.2}{0.3} & \fewshot{78.5}{0.3} & \fewshot{80.7}{0.3} & \fewshot{94.8}{0.2} & \fewshot{97.4}{0.1} & \fewshot{83.2}{0.2} & \fewshot{82.2}{0.3} & \fewshot{88.7}{0.2} & 78.2\\
LGSimCLR                       & RedCaps              & SBERT - Sub                    & \fewshot{88.7}{0.2} & \fewshot{70.4}{0.3} & \fewshot{78.4}{0.3} & \fewshot{75.3}{0.3} & \fewshot{90.7}{0.2} & \fewshot{67.3}{0.3} & \fewshot{47.6}{0.4} & \fewshot{78.3}{0.3} & \fewshot{76.6}{0.3} & \fewshot{95.4}{0.1} & \fewshot{97.8}{0.1} & \fewshot{86.5}{0.2} & \fewshot{85.3}{0.2} & \fewshot{89.1}{0.2} & 80.5\\
\midrule \rowheader \multicolumn{18}{l}{\textit{\textbf{Pre-training Datasets}}} \\
LGSimCLR                       & CC3M                 & SBERT (MPNet)                  & \fewshot{69.2}{0.3} & \fewshot{60.6}{0.3} & \fewshot{71.7}{0.3} & \fewshot{72.0}{0.3} & \fewshot{92.8}{0.2} & \fewshot{58.8}{0.3} & \fewshot{48.9}{0.4} & \fewshot{77.4}{0.3} & \fewshot{77.8}{0.3} & \fewshot{92.9}{0.2} & \fewshot{95.0}{0.2} & \fewshot{83.0}{0.3} & \fewshot{82.4}{0.3} & \fewshot{86.3}{0.3} & 76.3\\
LGSimCLR                       & CC12M                & SBERT (MPNet)                  & \fewshot{79.6}{0.3} & \fewshot{72.0}{0.3} & \fewshot{78.5}{0.3} & \fewshot{71.2}{0.3} & \fewshot{95.2}{0.1} & \fewshot{78.2}{0.3} & \fewshot{55.5}{0.4} & \fewshot{81.8}{0.3} & \fewshot{82.1}{0.3} & \fewshot{96.2}{0.1} & \fewshot{95.3}{0.1} & \fewshot{90.0}{0.2} & \fewshot{83.6}{0.3} & \fewshot{88.0}{0.2} & 81.9\\
LGSimCLR                       & RedCaps 2020         & SBERT (MPNet)                  & \fewshot{86.0}{0.2} & \fewshot{60.3}{0.3} & \fewshot{70.5}{0.3} & \fewshot{79.9}{0.3} & \fewshot{90.1}{0.2} & \fewshot{69.8}{0.3} & \fewshot{48.6}{0.4} & \fewshot{77.0}{0.3} & \fewshot{81.0}{0.3} & \fewshot{92.3}{0.2} & \fewshot{96.8}{0.1} & \fewshot{78.8}{0.3} & \fewshot{84.7}{0.2} & \fewshot{87.0}{0.2} & 78.8\\
\midrule \rowheader \multicolumn{18}{l}{\textit{\textbf{Batch Size Scaling}}} \\
SimCLR (256)                   & RedCaps              & -                              & \fewshot{64.9}{0.3} & \fewshot{52.7}{0.3} & \fewshot{57.9}{0.3} & \fewshot{51.4}{0.4} & \fewshot{86.6}{0.2} & \fewshot{43.6}{0.3} & \fewshot{38.3}{0.3} & \fewshot{77.1}{0.3} & \fewshot{68.7}{0.3} & \fewshot{79.2}{0.3} & \fewshot{85.9}{0.3} & \fewshot{69.7}{0.3} & \fewshot{84.1}{0.3} & \fewshot{81.3}{0.3} & 67.2\\
SimCLR (1024)                  & RedCaps              & -                              & \fewshot{67.5}{0.3} & \fewshot{54.0}{0.3} & \fewshot{59.2}{0.3} & \fewshot{53.0}{0.4} & \fewshot{87.2}{0.2} & \fewshot{44.7}{0.3} & \fewshot{38.9}{0.3} & \fewshot{77.9}{0.3} & \fewshot{71.5}{0.3} & \fewshot{80.4}{0.3} & \fewshot{87.9}{0.2} & \fewshot{71.8}{0.3} & \fewshot{82.5}{0.3} & \fewshot{81.8}{0.3} & 68.4\\
SimCLR (2048)                  & RedCaps              & -                              & \fewshot{68.4}{0.3} & \fewshot{51.8}{0.3} & \fewshot{57.8}{0.3} & \fewshot{53.3}{0.4} & \fewshot{87.3}{0.2} & \fewshot{45.4}{0.3} & \fewshot{38.8}{0.3} & \fewshot{77.8}{0.3} & \fewshot{73.2}{0.3} & \fewshot{81.6}{0.3} & \fewshot{87.9}{0.2} & \fewshot{71.6}{0.3} & \fewshot{84.0}{0.3} & \fewshot{81.6}{0.3} & 68.6\\
LGSimCLR (256)                 & RedCaps              & SBERT (MPNet)                  & \fewshot{90.3}{0.2} & \fewshot{66.6}{0.3} & \fewshot{75.8}{0.3} & \fewshot{81.6}{0.3} & \fewshot{92.6}{0.2} & \fewshot{75.3}{0.3} & \fewshot{50.5}{0.4} & \fewshot{81.6}{0.3} & \fewshot{83.2}{0.3} & \fewshot{95.3}{0.1} & \fewshot{97.6}{0.1} & \fewshot{86.8}{0.2} & \fewshot{86.4}{0.2} & \fewshot{88.9}{0.2} & 82.3\\
LGSimCLR (1024)                & RedCaps              & SBERT (MPNet)                  & \fewshot{90.3}{0.2} & \fewshot{64.9}{0.3} & \fewshot{75.7}{0.3} & \fewshot{83.8}{0.3} & \fewshot{92.6}{0.2} & \fewshot{78.2}{0.3} & \fewshot{52.6}{0.4} & \fewshot{80.9}{0.3} & \fewshot{83.6}{0.3} & \fewshot{95.6}{0.1} & \fewshot{97.6}{0.1} & \fewshot{86.7}{0.2} & \fewshot{86.0}{0.2} & \fewshot{88.6}{0.2} & 82.6\\
LGSimCLR (2048)                & RedCaps              & SBERT (MPNet)                  & \fewshot{90.6}{0.2} & \fewshot{67.5}{0.3} & \fewshot{76.6}{0.3} & \fewshot{83.9}{0.3} & \fewshot{92.6}{0.2} & \fewshot{79.7}{0.3} & \fewshot{51.5}{0.4} & \fewshot{80.6}{0.3} & \fewshot{83.8}{0.3} & \fewshot{95.8}{0.1} & \fewshot{97.6}{0.1} & \fewshot{87.1}{0.2} & \fewshot{86.6}{0.2} & \fewshot{89.2}{0.2} & 83.1\\
\midrule \rowheader \multicolumn{18}{l}{\textit{\textbf{Alternative Formulations}}} \\
LGSimCLR                       & RedCaps              & SBERT (MPNet)                  & \fewshot{90.3}{0.2} & \fewshot{66.3}{0.3} & \fewshot{75.5}{0.3} & \fewshot{83.1}{0.3} & \fewshot{92.7}{0.2} & \fewshot{77.6}{0.3} & \fewshot{50.6}{0.4} & \fewshot{81.1}{0.3} & \fewshot{84.1}{0.3} & \fewshot{95.4}{0.1} & \fewshot{97.6}{0.1} & \fewshot{86.5}{0.2} & \fewshot{85.0}{0.2} & \fewshot{89.0}{0.2} & 82.5\\
LGSimSiam                      & RedCaps              & SBERT (MPNet)                  & \fewshot{81.2}{0.3} & \fewshot{61.6}{0.3} & \fewshot{71.2}{0.3} & \fewshot{63.1}{0.4} & \fewshot{90.2}{0.2} & \fewshot{60.9}{0.3} & \fewshot{44.6}{0.4} & \fewshot{78.8}{0.3} & \fewshot{68.0}{0.3} & \fewshot{92.8}{0.2} & \fewshot{93.7}{0.2} & \fewshot{81.2}{0.3} & \fewshot{85.1}{0.2} & \fewshot{86.7}{0.2} & 75.7\\
LGSLIP                         & RedCaps              & SBERT (MPNet)                  & \fewshot{91.3}{0.2} & \fewshot{67.2}{0.3} & \fewshot{77.2}{0.3} & \fewshot{81.8}{0.3} & \fewshot{92.6}{0.2} & \fewshot{77.3}{0.3} & \fewshot{50.4}{0.4} & \fewshot{81.8}{0.3} & \fewshot{81.8}{0.3} & \fewshot{96.1}{0.1} & \fewshot{97.8}{0.1} & \fewshot{89.2}{0.2} & \fewshot{85.3}{0.2} & \fewshot{89.1}{0.2} & 82.8\\
\bottomrule
\end{tabular}
}
\end{table*}
\begin{table*}
  \newcommand{\rowheader}{\rowcolor{Gray!20}}
  \centering
    \caption{
  \textbf{ImageNet Evaluations.}
  We evaluate all models on several ImageNet robustness benchmarks.
  All models were trained using the ImageNet train set.
  We report the linear probe and few-shot classification performance. Subscripts show the 95\% confidence interval across 5000 episodes. }
  \label{tab:app_imageneteval}
  \setlength\tabcolsep{4pt}
  \footnotesize
% \resizebox{\linewidth}{!}{%
  \begin{tabularx}{\linewidth}{Xlc cccccccccc}
    \toprule
    & &
    & \multicolumn{5}{c}{Linear Probe}  
    & \multicolumn{5}{c}{Fewshot Classification} 
    \\
    \cmidrule(lr){4-8} 
    \cmidrule(lr){9-13}
    Model & \textbf{Dataset} &
    \textbf{Sampling Space} &
    % imagenet, imagenet_a, imagenet_r, imagenet_v2, imagenet_sketch
    \rotatebox[origin=lb]{90}{\bf {ImageNet}} &
    \rotatebox[origin=lb]{90}{\bf {ImageNet A}} &
    \rotatebox[origin=lb]{90}{\bf {ImageNet R}} &
    \rotatebox[origin=lb]{90}{\bf {ImageNet V2}} &
    \rotatebox[origin=lb]{90}{\bf {ImageNet Sketch}} &    
    \rotatebox[origin=lb]{90}{\bf {ImageNet}} &
    \rotatebox[origin=lb]{90}{\bf {ImageNet A}} &
    \rotatebox[origin=lb]{90}{\bf {ImageNet R}} &
    \rotatebox[origin=lb]{90}{\bf {ImageNet V2}} &
    \rotatebox[origin=lb]{90}{\bf {ImageNet Sketch}} \\
    \midrule
\rowheader \multicolumn{13}{l}{\textit{\textbf{Pre-trained Checkpoints}}} \\
Supervised~\cite{wightman2021resnet}      & ImageNet     & -                           &  80.7 &   5.4 &  27.6 &  68.9 &  28.8 & \fewshot{97.4}{0.1} & \fewshot{51.6}{0.2} & \fewshot{61.4}{0.2} & \fewshot{94.5}{0.1} & \fewshot{64.8}{0.3} \\
SimSiam~\cite{chen2021exploring}          & ImageNet     & -                           &  30.8 &   1.0 &  10.6 &  25.0 &   9.9 & \fewshot{90.4}{0.2} & \fewshot{39.8}{0.2} & \fewshot{52.9}{0.2} & \fewshot{85.6}{0.2} & \fewshot{56.3}{0.3} \\
MoCo v3~\cite{chen2021mocov3}             & ImageNet     & -                           &  69.5 &   1.1 &  20.3 &  57.2 &  20.5 & \fewshot{91.1}{0.1} & \fewshot{36.0}{0.2} & \fewshot{52.4}{0.2} & \fewshot{85.8}{0.2} & \fewshot{55.2}{0.3} \\
SwAV~\cite{caron2020unsupervised}         & ImageNet     & -                           &  70.6 &   1.2 &  16.6 &  57.5 &  17.6 & \fewshot{91.2}{0.1} & \fewshot{41.7}{0.2} & \fewshot{46.6}{0.2} & \fewshot{86.2}{0.2} & \fewshot{49.3}{0.3} \\
SimCLR~\cite{chen2020big}                 & ImageNet     & -                           &  68.7 &   0.9 &  16.0 &  56.1 &  15.7 & \fewshot{90.2}{0.2} & \fewshot{36.2}{0.2} & \fewshot{43.7}{0.2} & \fewshot{84.8}{0.2} & \fewshot{44.9}{0.3} \\
CLIP~\cite{radford2021learning}           & CLIP (400M)  & -                           &  73.2 &   8.2 &  31.9 &  61.5 &  31.8 & \fewshot{95.5}{0.1} & \fewshot{68.0}{0.2} & \fewshot{69.0}{0.2} & \fewshot{93.2}{0.1} & \fewshot{74.5}{0.3} \\
\midrule \rowheader \multicolumn{13}{l}{\textit{\textbf{RedCaps-trained Baselines}}} \\
SwAV                           & RedCaps              & -                              &  52.1 &   0.8 &   7.0 &  39.0 &   6.6 & \fewshot{80.4}{0.2} & \fewshot{38.6}{0.2} & \fewshot{35.1}{0.2} & \fewshot{75.5}{0.2} & \fewshot{33.1}{0.2} \\
SimSiam                        & RedCaps              & -                              &  52.9 &   0.8 &   8.0 &  40.3 &   8.7 & \fewshot{78.8}{0.2} & \fewshot{39.1}{0.2} & \fewshot{38.9}{0.2} & \fewshot{73.4}{0.2} & \fewshot{39.3}{0.2} \\
SimCLR                         & RedCaps              & -                              &  56.2 &   0.8 &   8.4 &  42.4 &   8.9 & \fewshot{79.8}{0.2} & \fewshot{39.6}{0.2} & \fewshot{38.7}{0.2} & \fewshot{74.4}{0.2} & \fewshot{38.6}{0.2} \\
Visual NNCLR                   & RedCaps              & -                              &  54.4 &   0.8 &   8.3 &  41.0 &   8.3 & \fewshot{81.3}{0.2} & \fewshot{39.5}{0.2} & \fewshot{38.0}{0.2} & \fewshot{76.3}{0.2} & \fewshot{36.8}{0.2} \\
CLIP                           & RedCaps              & -                              &  62.6 &   2.1 &  14.5 &  49.8 &  13.7 & \fewshot{88.7}{0.2} & \fewshot{44.7}{0.2} & \fewshot{46.6}{0.2} & \fewshot{84.7}{0.2} & \fewshot{46.1}{0.3} \\
CLIP (SBERT Encoder)           & RedCaps              & -                              &  61.5 &   2.1 &  14.4 &  49.2 &  13.2 & \fewshot{89.1}{0.2} & \fewshot{42.6}{0.2} & \fewshot{46.4}{0.2} & \fewshot{84.7}{0.2} & \fewshot{49.2}{0.3} \\
Language NNCLR                 & RedCaps              & -                              &  61.6 &   2.1 &  13.7 &  49.6 &  13.3 & \fewshot{89.0}{0.2} & \fewshot{45.1}{0.2} & \fewshot{47.4}{0.2} & \fewshot{84.9}{0.2} & \fewshot{48.8}{0.3} \\
SLIP                           & RedCaps              & -                              &  62.6 &   0.9 &  12.6 &  49.2 &  12.5 & \fewshot{86.7}{0.2} & \fewshot{43.0}{0.2} & \fewshot{43.5}{0.2} & \fewshot{82.0}{0.2} & \fewshot{43.8}{0.3} \\
\midrule \rowheader \multicolumn{13}{l}{\textit{\textbf{Sampling Space - Language}}} \\
LGSimCLR                       & RedCaps              & SBERT (MiniLM)                 &  65.3 &   1.2 &  16.8 &  52.8 &  16.8 & \fewshot{91.0}{0.1} & \fewshot{45.1}{0.2} & \fewshot{58.4}{0.2} & \fewshot{86.6}{0.2} & \fewshot{60.8}{0.3} \\
LGSimCLR                       & RedCaps              & CLIP (400M)                    &  65.7 &   1.2 &  16.7 &  53.0 &  16.8 & \fewshot{90.9}{0.2} & \fewshot{45.4}{0.2} & \fewshot{57.1}{0.2} & \fewshot{86.6}{0.2} & \fewshot{59.2}{0.3} \\
LGSimCLR                       & RedCaps              & CLIP (RedCaps)                 &  65.4 &   1.3 &  16.9 &  53.0 &  16.5 & \fewshot{90.6}{0.2} & \fewshot{45.4}{0.2} & \fewshot{57.1}{0.2} & \fewshot{86.2}{0.2} & \fewshot{58.5}{0.3} \\
LGSimCLR                       & RedCaps              & FastText BoW                   &  62.6 &   0.6 &  15.5 &  49.7 &  14.9 & \fewshot{89.7}{0.2} & \fewshot{44.0}{0.2} & \fewshot{55.8}{0.2} & \fewshot{84.9}{0.2} & \fewshot{57.0}{0.3} \\
\midrule \rowheader \multicolumn{13}{l}{\textit{\textbf{Sampling Space - Visual}}} \\
LGSimCLR                       & RedCaps              & ImageNet Supervised            &  66.9 &   0.8 &  19.0 &  53.6 &  16.7 & \fewshot{91.2}{0.1} & \fewshot{41.4}{0.2} & \fewshot{56.0}{0.2} & \fewshot{86.1}{0.2} & \fewshot{53.4}{0.3} \\
LGSimCLR                       & RedCaps              & SimCLR                         &  63.0 &   0.8 &  12.8 &  49.1 &  12.2 & \fewshot{88.4}{0.2} & \fewshot{38.7}{0.2} & \fewshot{45.8}{0.2} & \fewshot{83.1}{0.2} & \fewshot{46.0}{0.3} \\
LGSimCLR                       & RedCaps              & CLIP (400M)                    &  68.4 &   1.3 &  18.1 &  55.2 &  16.7 & \fewshot{90.9}{0.2} & \fewshot{45.2}{0.2} & \fewshot{55.1}{0.2} & \fewshot{86.3}{0.2} & \fewshot{52.7}{0.3} \\
\midrule \rowheader \multicolumn{13}{l}{\textit{\textbf{Sampling Scope}}} \\
LGSimCLR                       & RedCaps              & SBERT - Year                   &  64.5 &   0.9 &  15.6 &  51.6 &  15.9 & \fewshot{88.9}{0.2} & \fewshot{43.2}{0.2} & \fewshot{53.2}{0.2} & \fewshot{84.2}{0.2} & \fewshot{51.8}{0.3} \\
LGSimCLR                       & RedCaps              & SBERT - Sub-Year               &  66.2 &   1.3 &  19.0 &  53.2 &  20.1 & \fewshot{85.8}{0.2} & \fewshot{41.2}{0.2} & \fewshot{53.3}{0.3} & \fewshot{80.6}{0.2} & \fewshot{53.1}{0.3} \\
LGSimCLR                       & RedCaps              & SBERT - Sub                    &  67.0 &   1.5 &  19.5 &  54.4 &  20.8 & \fewshot{80.1}{0.2} & \fewshot{38.8}{0.2} & \fewshot{48.2}{0.3} & \fewshot{75.3}{0.2} & \fewshot{42.9}{0.3} \\
\midrule \rowheader \multicolumn{13}{l}{\textit{\textbf{Pre-training Datasets}}} \\
LGSimCLR                       & CC3M                 & SBERT (MPNet)                  &  17.5 &   1.0 &   8.9 &  13.4 &   5.4 & \fewshot{87.9}{0.2} & \fewshot{41.2}{0.2} & \fewshot{56.6}{0.2} & \fewshot{83.5}{0.2} & \fewshot{60.8}{0.3} \\
LGSimCLR                       & CC12M                & SBERT (MPNet)                  &  65.0 &   0.7 &  22.8 &  52.1 &  27.1 & \fewshot{91.9}{0.1} & \fewshot{45.7}{0.2} & \fewshot{66.0}{0.2} & \fewshot{87.8}{0.2} & \fewshot{73.0}{0.3} \\
LGSimCLR                       & RedCaps 2020         & SBERT (MPNet)                  &  58.7 &   0.6 &  13.1 &  45.4 &  11.4 & \fewshot{87.5}{0.2} & \fewshot{41.1}{0.2} & \fewshot{52.1}{0.2} & \fewshot{82.3}{0.2} & \fewshot{51.3}{0.3} \\
\midrule \rowheader \multicolumn{13}{l}{\textit{\textbf{Batch Size Scaling}}} \\
SimCLR (256)                   & RedCaps              & -                              &  54.8 &   0.7 &   7.9 &  41.6 &   8.1 & \fewshot{79.5}{0.2} & \fewshot{39.5}{0.2} & \fewshot{38.4}{0.2} & \fewshot{74.0}{0.2} & \fewshot{38.3}{0.2} \\
SimCLR (1024)                  & RedCaps              & -                              &  57.2 &   0.7 &   8.8 &  43.7 &   8.7 & \fewshot{80.5}{0.2} & \fewshot{39.4}{0.2} & \fewshot{39.2}{0.2} & \fewshot{75.1}{0.2} & \fewshot{38.3}{0.2} \\
SimCLR (2048)                  & RedCaps              & -                              &  58.1 &   0.8 &   9.0 &  44.5 &   9.2 & \fewshot{81.1}{0.2} & \fewshot{39.3}{0.2} & \fewshot{39.5}{0.2} & \fewshot{75.8}{0.2} & \fewshot{38.9}{0.2} \\
LGSimCLR (256)                 & RedCaps              & SBERT (MPNet)                  &  64.9 &   1.2 &  16.7 &  52.0 &  16.6 & \fewshot{90.6}{0.2} & \fewshot{46.3}{0.2} & \fewshot{57.4}{0.2} & \fewshot{86.3}{0.2} & \fewshot{59.7}{0.3} \\
LGSimCLR (1024)                & RedCaps              & SBERT (MPNet)                  &  65.8 &   1.2 &  16.7 &  52.9 &  16.8 & \fewshot{90.7}{0.2} & \fewshot{44.8}{0.2} & \fewshot{57.2}{0.2} & \fewshot{86.3}{0.2} & \fewshot{58.8}{0.3} \\
LGSimCLR (2048)                & RedCaps              & SBERT (MPNet)                  &  66.2 &   1.1 &  17.3 &  53.1 &  17.4 & \fewshot{90.8}{0.2} & \fewshot{44.7}{0.2} & \fewshot{58.0}{0.2} & \fewshot{86.3}{0.2} & \fewshot{60.1}{0.3} \\
\midrule \rowheader \multicolumn{13}{l}{\textit{\textbf{Alternative Formulations}}} \\
LGSimCLR                       & RedCaps              & SBERT (MPNet)                  &  65.2 &   1.1 &  16.6 &  52.5 &  16.2 & \fewshot{90.9}{0.2} & \fewshot{45.0}{0.2} & \fewshot{57.4}{0.2} & \fewshot{86.4}{0.2} & \fewshot{59.2}{0.3} \\
LGSimSiam                      & RedCaps              & SBERT (MPNet)                  &  58.9 &   0.7 &  12.2 &  45.9 &  12.2 & \fewshot{88.1}{0.2} & \fewshot{44.1}{0.2} & \fewshot{48.8}{0.2} & \fewshot{83.5}{0.2} & \fewshot{50.7}{0.3} \\
LGSLIP                         & RedCaps              & SBERT (MPNet)                  &  66.8 &   1.4 &  18.2 &  54.3 &  18.9 & \fewshot{90.4}{0.2} & \fewshot{46.2}{0.2} & \fewshot{58.4}{0.2} & \fewshot{86.2}{0.2} & \fewshot{59.7}{0.3} \\
\bottomrule
\end{tabularx}
% }
\end{table*}

%%%%%%%%% REFERENCES

\end{document}